\documentclass[a4paper,fleqn]{cas-dc}
\usepackage[numbers,sort&compress]{natbib}
\usepackage{multirow}
\usepackage{dsfont}
\usepackage{placeins} % 用于控制浮动体的排版
\usepackage[pagewise]{lineno}
\usepackage{pifont}
\usepackage[numbers]{natbib}

\def\tsc#1{\csdef{#1}{\textsc{\lowercase{#1}}\xspace}}
\tsc{WGM}
\tsc{QE}
\tsc{EP}
\tsc{PMS}
\tsc{BEC}
\tsc{DE}
\usepackage{tabularx}

\begin{document}
%Short title
\shorttitle{DPG-CD}
\shortauthors{Zhang et~al.}

\let\WriteBookmarks\relax
\def\floatpagepagefraction{1}
\def\textpagefraction{.001}

\newcommand{\reviewed}[1]{\textcolor{blue}{#1}} 

%\usepackage{graphicx}

% Short title
\shorttitle{DPG-CD}
\shortauthors{Zhang et~al.}

% Main title of the paper
\title [mode = title]{DPG-CD: Depth-Prior-Guided Cross-Modal Joint 2D–3D Change Detection}                      
% Title footnote mark
% eg: \tnotemark[1]
%\tnotemark[1,2]

% Title footnote 1.
% eg: \tnotetext[1]{Title footnote text}
% \tnotetext[<tnote number>]{<tnote text>} 
%\tnotetext[1]{This document is the results of the research
%   project funded by the National Science Foundation.}

%\tnotetext[2]{The second title footnote which is a longer text matter
%   to fill through the whole text width and overflows into
%   another line in the footnotes area of the first page.}

% First author
%
% Options: Use if required
% eg: \author[1,3]{Author Name}[type=editor,
%       style=chinese,
%       auid=000,
%       bioid=1,
%       prefix=Sir,
%       orcid=0000-0000-0000-0000,
%       facebook=<facebook id>,
%       twitter=<twitter id>,
%       linkedin=<linkedin id>,
%       gplus=<gplus id>]
\author[1]{Luqi Zhang}
% [type=editor,
                        % auid=000,bioid=1,
                        % prefix=Sir,
                        % role=Researcher,
%                         orcid=0000-0001-7511-2910]

% Corresponding author indication

% Footnote of the first author
% \fnmark[1]

% Email id of the first author
%\ead{martin_liao@whu.edu.cn}

% URL of the first author
% \ead[url]{www.cvr.cc, cvr@sayahna.org}

%  Credit authorship
% \credit{Conceptualization of this study, Methodology, Software}

% Address/affiliation
\affiliation[1]{organization={State Key Laboratory of Information Engineering in Surveying, Mapping and Remote Sensing, Wuhan University},
    addressline={No.129, Luoyu Road}, 
    city={Wuhan},
    state={Hubei},
    postcode={430079}, 
    country={China}}

\author[1]{Zhen Dong}
\author[1]{Bisheng Yang}
\cormark[1]

% Corresponding author text
% \cortext[cor1]{Equal Contribution}
\cortext[cor2]{Corresponding author}
% \tnotetext[1]{This research was supported by the National Natural Science Foundation Project (No.42130105)} 

% Footnote text
%\fntext[fn1]{This is the first author footnote. but is common to third
%  author as well.}
%\fntext[fn2]{Another author footnote, this is a very long footnote and
%  it should be a really long footnote. But this footnote is not yet
%  sufficiently long enough to make two lines of footnote text.}

% For a title note without a number/mark
%\nonumnote{This note has no numbers. In this work we demonstrate $a_b$
%  the formation Y\_1 of a new type of polariton on the interface
%  between a cuprous oxide slab and a polystyrene micro-sphere placed
%  on the slab.
%  }
\renewcommand{\arraystretch}{1.05}
% Here goes the abstract
\begin{abstract}
Urban spatial evolution is manifested not only through horizontal expansion but also through vertical structural changes. Consequently, jointly capturing 2D semantic changes and 3D height changes is essential for urban morphology analysis and emergency management.
In practical scenarios, collecting 3D observations is often constrained by high acquisition costs and the inability to support frequent updates. The multi-temporal cross-modal  input consisting of pre-event Digital Surface Model (DSM) and post-event imagery provides a practical solution for 3D change detection in high-frequency urban monitoring, disaster assessment, and emergency response scenarios.
However, this setting remains challenging as imagery and DSM data exhibit significant spectral–geometric representation gaps. Moreover, modality differences may be confused with actual changes, and robust change detection requires effective fusion of semantic and geometric features from multi-temporal data.
In this paper, we propose DPG-CD, a depth-prior-guided multi-temporal cross-modal fusion framework for joint 2D semantic and 3D height change detection. 
Specifically, an estimated depth prior is introduced into the imagery to mitigate the modality gap with DSM. A gated fusion mechanism then selectively injects geometric cues from depth prior while preserving discriminative spectral representations. Subsequently, a multi-stage cross-temporal cross-modal feature fusion architecture is employed to extract change-aware features. Finally, a multi-task decoder jointly predicts 2D semantic changes and 3D height changes, complemented by an auxiliary DSM prediction task to improve structural consistency and height estimation accuracy.
Experiments on two public datasets, Hi-BCD and 3DCD, and a new dataset, NYC-MMCD, demonstrate that DPG-CD outperforms state-of-the-art methods on both 2D and 3D change detection tasks. The code and the NYC-MMCD dataset will be made publicly available online at:  \url{https://github.com/zhangluqi0209/DPG-CD}.

\end{abstract}

% Use if graphical abstract is present
% \begin{graphicalabstract}
% \includegraphics{figs/grabs.pdf}
% \end{graphicalabstract}

% Research highlights
% \begin{highlights}
% \item This paper proposes a multi-temporal cross-modal fusion-based change detection method that leverages pre-event DSM and post-event optical imagery, and employs depth prior guidance to mitigate the significant modality gap across multi-temporal data.
% \item A multi-stage cross-modal feature fusion framework is constructed to enable effective extraction of change features across different epochs.
% \item It jointly predicts 2D semantic changes and 3D height changes, while using DSM prediction corresponding to the image modality as an auxiliary task to improve the accuracy of 3D height change estimation.
% \item A new cross-modal change detection dataset, NYC-MMCD, is introduced for multi-temporal cross-modal joint 2D-3D change detection.
% \end{highlights}

% Keywords
% Each keyword is seperated by \sep
\begin{keywords}
Cross-modal Change Detection \sep 
3D Change Detection\sep  
Estimated Depth Prior \sep
Multi-task Learning
\end{keywords}

\maketitle

\section{Introduction}
% \linenumbers
% \pagewiselinenumbers
% \switchlinenumbers

The regular utilization of multi-source remote sensing data for detecting Earth’s surface changes is pivotal across a wide range of applications, including land-use change analysis \citep{piao2021analysis}, building change detection \citep{zhou2022machine, argialas2016building}, disaster response \citep{wang2026few}, and urban vegetation monitoring \citep{wang2022novel}.
Driven by the increasing diversity of acquisition platforms and sensing modalities, change detection has gradually evolved from a single-source comparison to multi-source information fusion \citep{bai2023deep, zhang2021novel}.
Integrating multi-source spatiotemporal urban data is crucial for holistic urban perception, dynamic analysis, and intelligent governance \citep{dong2026neural}.
Based on the types of multi-temporal data employed, change detection methods can be categorized into homogeneous and heterogeneous approaches. Compared to homogeneous data, multi-modal data are not constrained by the requirement for observations from the same sensor or platform, thereby facilitating higher temporal resolution and offering superior flexibility in emergency scenarios \citep{jiang2024change_re, li2022deep}.

Furthermore, urban space is intrinsically 3D, and its development is increasingly characterized by intensive utilization of vertical directions \citep{zhang2025mapping}. While 3D change detection is indispensable for capturing vertical urban growth, existing literature on building changes remains predominantly confined to 2D representations \citep{papini2025evolving}.
3D change detection is a subset of remote sensing change detection. Incorporating 3D data with height information, it enables the extraction of volumetric and elevation differences \citep{qin20163d}. However, in practical applications, simultaneously acquiring multi-temporal 3D data alongside corresponding imagery leads to costs in both data acquisition and computational processing.

In practice, large-scale historical 3D data recorded before an event are often readily accessible, while the acquisition of timely post-event 3D data remains challenging, especially for high-frequency monitoring. In recent years, driven by the need for flexible monitoring and the increasing demand for 3D change estimation, combining 2D imagery with 3D data for joint 2D-3D change detection has gained significant traction in recent years.
Among various modality combinations, a multi-temporal cross-modal fusion of pre-event 3D DSM data and post-event 2D imagery is particularly compelling. While optical imagery provides rich spectral and semantic information, DSM offer explicit height and structural geometry, enabling a more complete characterization of building morphology and urban forms \citep{mutreja2025}.
 
Despite its potential, multi-temporal cross-modal fusion for joint 2D-3D change detection poses significant challenges. First, DSM and imagery data exhibit disparate in statistical distributions, resulting in the spectral-geometric representation gap that complicates effective cross-modal alignment \citep{chen2023fourier}. Second, cross-modality discrepancies are often confounded with actual changes, thereby impeding reliable change-aware feature extraction from heterogeneous signals \citep{zhu2023r2fd2, jiang2024change}. Third, jointly modeling 2D semantic changes and 3D height changes within a unified framework requires a cohesive fusion strategy across semantic and geometric features.

To tackle these issues, this paper introduces DPG-CD, a depth-prior-guided cross-modal fusion framework designed for flexible and high-resolution urban change detection. Leveraging estimated depth prior and employing a gated fusion mechanism bridge the representation gap between imagery and DSM while selectively injecting geometric cues into the optical modality. Then, the proposed method synergistically predicts 2D semantic changes and 3D height changes within a multi-stage cross-modal feature fusion architecture. Furthermore, supervision from the DSM corresponding to the image modality is incorporated to further enhance cross-modal alignment and change detection accuracy.

The main contributions of this paper are as follows:

(1) To reduce the gap between bi-temporal data from different modalities, we introduce an estimated depth prior from a monocular depth estimation foundation model into the image branch. A gated fusion mechanism is then used to combine the depth prior features with the spectral features of the image. This design helps reduce modality differences and improves feature consistency between the bi-temporal inputs.

(2) To better capture cross-temporal change features, we develop a multi-stage cross-modal  feature fusion architecture. At different feature levels, a convolutional channel attention block (CCAB) and a hierarchical change feature extraction block (HCFEB) are used to model interactions between heterogeneous modalities across two temporal phases. These modules strengthen the representation of change-related features.

(3) To support robust multi-temporal cross-modal fusion, we design a multi-task decoder with auxiliary DSM prediction to jointly optimize semantic and geometric outputs. This design improves structural consistency and enhances the accuracy of joint 2D-3D change detection.

(4) To support rigorous evaluation and advance method development for multi-temporal cross-modal fusion-based in joint 2D-3D change detection, a new multi-modal dataset, NYC-MMCD, is introduced. The dataset provides multi-class 2D semantic building changes, including newly built and demolition categories. It also contains complex 3D height change samples across diverse urban morphologies.

The remainder of this paper is organized as follows. Section \ref{sec_related_work} reviews related work on multi-modal change detection and 3D change detection. Section \ref{sec_method} presents the proposed DPG-CD framework in detail. Section \ref{sec_exp_settings} describes the evaluation metrics, datasets and implementation details. Section \ref{sec_exp_re} reports and analyzes the experimental results. Section \ref{sec_dicussion} presents the ablation studies and parameter sensitivity analysis. Finally, Section \ref{sec_conclusion} concludes the paper and discusses future work.

\section{Related works \label{sec_related_work}}
\subsection{Change Detection with Multi-modal Data}
Multi-modal remote sensing data offer complementary observations of Earth's surface characteristics. By synergistically leveraging multi-source data, the interpretation of ground objects can be significantly enhanced, leading to robust performance improvements in tasks such as semantic segmentation, depth estimation, and change detection \citep{wang2025mssdf}. Multi-modal change detection methods typically take multi-modal data as input at each temporal phase, allowing complementary information from different modalities to be fused.

In change detection, multi-temporal optical imagery is often affected by spectral confusion, occlusions, and viewpoint variations. Consequently, an increasing number of studies have integrated auxiliary modalities to bolster change detection performance. For example, DSM data can directly represent the geometric structure of ground objects, serving as a vital complement to 2D spectral cues. The height information provided by DSM effectively alleviates the ambiguity in traditional 2D change detection, thereby enhancing the precision of structural delineation \citep{you2020survey}. Accordingly, the synergistic utilization of optical imagery and DSM across bi-temporal phases has emerged as a pivotal research direction.

A common strategy involves the simultaneous utilization of optical imagery and DSM data from two epochs, and predict 2D and 3D changes separately. For instance, \citet{singla2023two} conducts joint modeling of horizontal and vertical changes to facilitate a comprehensive assessment of urban dynamics.
Similarly, \citet{quan2025change} develops a 3D change detection framework for GF-7 satellite imagery that integrates DSM and optical imagery through feature alignment and cross-modal attention. This approach enables fine-grained identification of change boundaries in complex mining scenarios and allows for the estimation of excavation volumes based on elevation difference integration.
Additionally, \citet{fu2025dddmnet} incorporates multi-source data, including DSM, differential DSM, DOM, and NDVI, to model change information from geometric, spectral, and environmental perspectives. 
%Among these modalities, differential DSM demonstrate distinct advantages in capturing vertical structural changes.

Traditional methods mainly relied on handcrafted features and probabilistic graphical models for multi-modal fusion. For instance, \citet{tian2013building} integrates DSM derived from stereo matching with spectral imagery, using height information to distinguish spectrally similar objects. \citet{pang2019co} fuses point clouds and imagery at the supervoxel level and performs change detection as an energy optimization problem. \citet{wang2022building} constructs spectral and height change indicators and applies a graph-cut model to achieve spatially coherent change segmentation.
Furthermore, \citet{tian2021three} employs evidence theory to facilitate the complementary fusion of DSM and NDVI difference information. By using basic probability assignment and evidence combination rules, these approach alleviates conflicts between heterogeneous data and improves robustness in complex scenarios.

With the advancement of deep learning, multi-modal change detection has gradually transitioned toward end-to-end feature extraction frameworks that utilize various fusion strategies to bolster detection accuracy.
For example, early fusion strategies directly concatenate multi-temporal RGB imagery and DSM at the input layer, thereby facilitating the learning of unified feature representations \citep{gomroki2023automatic}. 
Alternatively, intermediate feature fusion often involves modality-specific encoders to extract optical and geometric features independently. Subsequently, self-attention mechanisms are integrated into deeper layers to model the complementary relationships between spectral and geometric information \citep{pan2022self}.
Moreover, recent studies explore Transformer-based frameworks that extract multi-source information during encoding and perform cross-modal interaction in a unified feature space. This integration effectively enhances the discriminative capacity for complex change patterns \citep{teo2025building}.

In summary, existing multi-modal change detection methods improve performance by jointly leveraging complementary information across different modalities at each temporal phase. However, these approaches typically presume that coincident multi-modal observations are available at each epoch. This assumption is often difficult to satisfy in large-scale or time-sensitive scenarios due to data acquisition constraints. These limitations motivate further study of more challenging multi-temporal cross-modal settings, where reliable historical 3D data and timely imagery can be fused under incomplete multi-temporal observations.

\subsection{3D Change Detection with Cross-modal Data}

3D change detection plays a pivotal role in rapid urban monitoring, disaster assessment, and infrastructure management because it can captures structural evolution from a geometric perspective. In practical applications, using different modalities for the two temporal phase enables more flexible implementation of 3D change detection. In this setting, reliable historical 3D model is often available, while subsequent observations are primarily derived from optical imagery. Compared to traditional 2D change detection, this paradigm is particularly well-suited for large-scale and time-sensitive scenarios, as it facilitates the inference of 3D changes without necessitating the frequent and costly re-acquisition of 3D data.

Early studies demonstrated the concept of cross-modal change detection by combining archival 3D reference data with subsequent imagery. Compared to airborne LiDAR, stereo or aerial imagery can provide large-scale urban 3D structural information at a lower cost.
%and has therefore been widely used in related studies.
Based on stereo geometry principles, existing 3D geographic models can serve as reference, where changes are detected by evaluating geometric consistency \citep{qin20163d}. For example, \citet{zhou2020lidar} converts existing LiDAR point clouds into candidate DSMs to constrain the disparity search space in the dense matching of subsequent stereo imagery, thereby facilitating change detection through spectral discrepancies. \citet{qin2014change} directly performs change detection on 3D building models by rasterizing the models and generating DSMs from stereo imagery. The authors further integrate geometric consistency, elevation differences, and texture similarity to formulate change indicators for building change detection and model updating. While these methods reduce the need for repeated LiDAR acquisition, their performance is still highly dependent on stereo reconstruction quality and is easily affected by weak textures, occlusions, and illumination variations.

To further simplify data acquisition, several studies have explored the possibility of bypassing explicit 3D data entirely. For example, recent research has moved toward directly inferring 3D changes solely from imagery, without requiring explicit 3D observations. For instance, \citet{marsocci2023inferring} proposes a multi-task framework that takes only bi-temporal optical images as input, jointly modeling 2D and 3D changes through a Transformer-based encoder and a task decoupling mechanism, without requiring explicit elevation data. Similarly, \citet{wang2026dsti} introduces the DSTI-Net multi-task framework designed to jointly models 2D semantics and 3D height changes by leveraging Segment Anything Model to guide change-aware feature extraction. Furthermore, \citet{meng2025changeda} implicitly learns geometric information from imagery using a deep encoder and enhances 3D structural change perception through a depth injection module, enabling joint prediction of 2D and 3D changes. While these methods offer enhanced flexibility, they do not explicitly leverage reliable historical 3D observations, leading to unverifiable generalization performance.

More recently, research has progressed towards multi-temporal cross-modal fusion by jointly exploiting reliable 3D data and optical imagery for 3D change detection. For example, \citet{liu2024transformer} extracts shared DSM-image representations through cross-modal interactive attention, incorporating consistency constraints to mitigate conflicts between semantic and height change prediction. Moreover, \citet{liu2025hatformer} proposes a height-aware Transformer framework to tackle cross-modal discrepancies and foreground-background imbalance. This method explicitly models interactions between optical and DSM features and utilizes background supervision to support joint 2D and 3D change detection. Although these studies demonstrate the effectiveness of integrating historical 3D geometric cues with contemporary imagery, several critical challenges persist. First, bridging the representation gap between heterogeneous modalities remains a significant hurdle. Furthermore, accurately disentangling inherent modality discrepancies from actual changes remains insufficiently addressed.

\section{DPG-CD \label{sec_method}}
\begin{figure*}[h]
\centering
\includegraphics[width=0.8\linewidth]{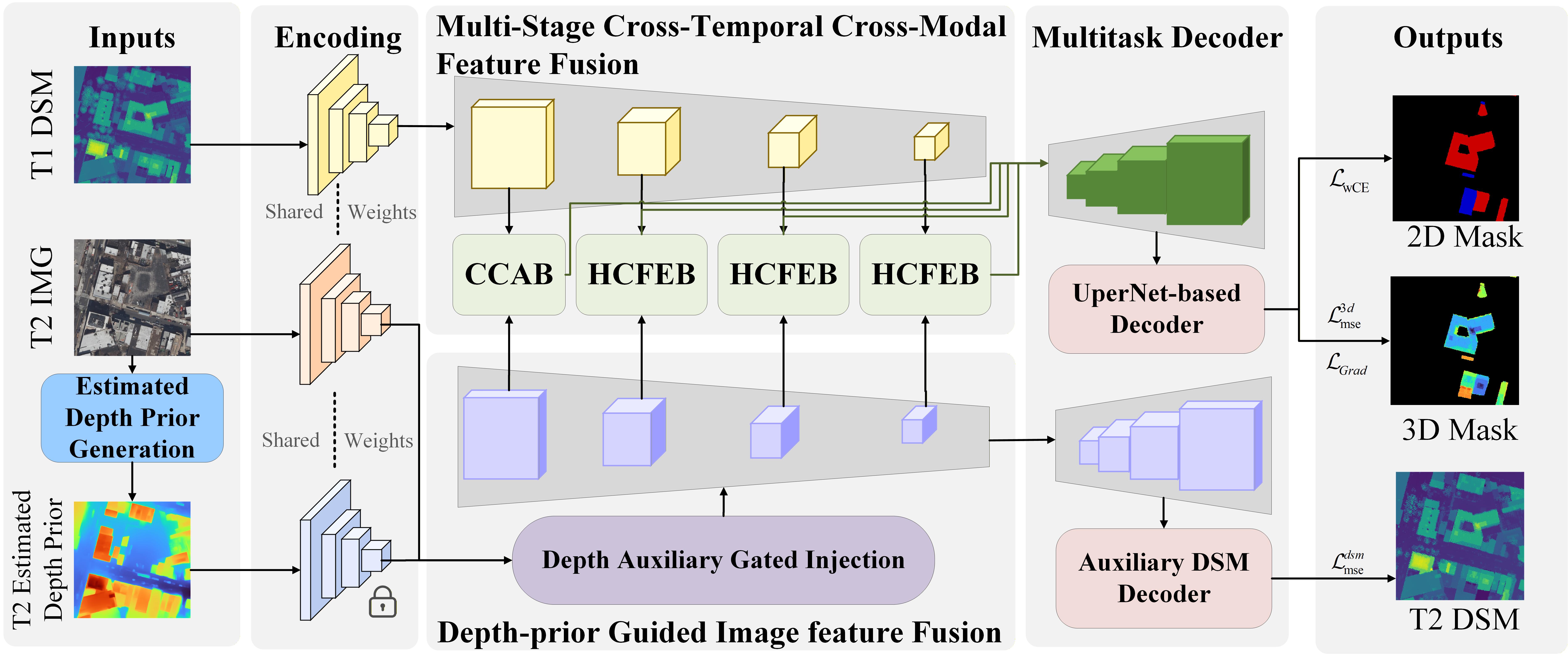}
\caption{The overall architecture of the proposed Depth-Prior-Guided framework for joint 2D–3D change detection}
\label{fig:framework}
\end{figure*}

To address change detection with cross-modal data, this paper proposes DPG-CD, a depth-prior-guided cross-modal change detection method, which jointly predict 2D semantic changes and 3D height changes. DPG-CD first introduces an estimated depth prior for the image modality. 
Then, a gated mechanism is employed to fuse image features with depth prior features, enhancing the geometric representation of the image modality and improving its consistency with the DSM modality.
Subsequently, a multi-stage cross-temporal cross-model fusion architecture is adopted to extract change-related features. 
Finally, a multi-task decoding scheme is employed to separately predict 2D semantic changes and 3D height changes, with the DSM prediction from imagery serving as an auxiliary task.

\subsection{Multi-Modal Feature Extraction}

\subsubsection{Estimated Depth Prior Generation}
To enable higher-frequency and more flexible 3D change detection, DPG-CD leverages 3D DSM data together with high-resolution 2D aerial imagery.
High-resolution imagery captures fine-grained texture of ground objects but lacks explicit 3D geometric information, whereas DSM is a raster-based representation of surface elevation that reflects the distribution of urban objects. Due to the significant modality gap between image features and DSM features, this method introduces estimated depth prior to bridge the discrepancy. 
%The proposed method further exploits geometric cues embedded in imagery, enhancing the alignment between image and DSM modalities, thereby enabling more accurate extraction of change-related features during decoding.

Large-scale pre-trained models learned from massive datasets can provide rich prior knowledge for image understanding. In this work, we leverage the depth foundation model Depth Anything V2 \citep{NEURIPS2024_26cfdcd8} as a complement to the geometric modality corresponding to imagery.
Depth Anything V2 is a monocular depth model that estimates relative depth from a single RGB image. When applied to aerial imagery, the estimated depth reflects elevation distribution of the ground and buildings. In this paper, given the T2 aerial image $\mathbf{I}^{img}$
, the pre-trained model is directly applied to generate the corresponding estimated depth map 
$\mathbf{D}_{\mathrm{edm}}$ as the depth prior.
\begin{equation}\label{eq:dpv2}
 \mathbf{D}_{\mathrm{edm}} = \Phi_{\theta^{\ast}}(\mathbf{I}^{img} )
 \end{equation}
where, 
$\Phi_{\theta^{\ast}}()$ denotes the Depth Anything V2 model with pre-trained weights 
$\theta^{\ast}$.

\subsubsection{MambaVision Backbone for Feature Extraction}
DPG-CD adopts the MambaVision backbone for feature extraction \citep{Hatamizadeh_2025_CVPR}, which is used to obtain multi-level features from the input images. 
%MambaVision extracts features across four stages: the first two stages employ CNN-based layers, while the third and fourth stages utilize MambaVision and Transformer blocks for higher-level semantic modeling.
Each input image is processed by the MambaVision encoder to generate four-level hierarchical features. 
%The structure and outputs of the encoder is illustrated in the Fig.~\ref{fig:encoder}.

Following \citet{liu2024transformer}, the single-channel DSM modality is normalized based on its input values and converted into a grayscale image. The image modality and DSM modality are then fed into a backbone with shared weights $\mathcal{E}$ to extract hierarchical multi-level features. 
%In this work, the backbone outputs four levels of features for each modality.
\begin{equation}\label{eq:encoder}
\begin{aligned}
\left\{ \mathbf{F}_{1}^{dsm}, \mathbf{F}_{2}^{dsm}, \mathbf{F}_{3}^{dsm}, \mathbf{F}_{4}^{dsm} \right\}
&= \mathcal{E}\!\left( \mathbf{I}^{dsm} \right), \\
\left\{ \mathbf{F}_{1}^{img}, \mathbf{F}_{2}^{img}, \mathbf{F}_{3}^{img}, \mathbf{F}_{4}^{img} \right\}
&= \mathcal{E}\!\left( \mathbf{I}^{img} \right).
\end{aligned}
\end{equation}
where, $\mathcal{E}$ denotes the MambaVision backbone for multi-level feature extraction.

The estimated depth map $\mathbf{D}_{\mathrm{edm}}$ and the DSM both represent elevation variations of urban objects, serving similar roles in capturing surface structure and spatial distribution.
In this work, the estimated depth map 
 is processed in the same manner as the DSM modality, it is normalized and converted into an grayscale image, and then fed into the encoder to extract multi-level features.
As an auxiliary input, the estimated depth map is kept frozen within the feature extraction encoder.

\begin{equation}\label{eq:est}
\left\{ \mathbf{F}_{1}^{edm}, \mathbf{F}_{2}^{edm}, \mathbf{F}_{3}^{edm}, \mathbf{F}_{4}^{edm} \right\}
= \mathcal{E}_{\mathrm{freeze}}\!\left( \mathbf{D}_{\mathrm{edm}}  \right).
 \end{equation}

% \begin{figure*}[h]
% \centering
% \includegraphics[width=\linewidth]{figs/encoder.jpg}
% \caption{The MambaVision-based hierarchical image feature extraction framework adopted in this paper.}
% \label{fig:encoder}
% \end{figure*}

\subsection{Depth-Prior-Guided Image feature Fusion}  \label{sec:Gated}
Considering that the estimated depth prior may contain errors and noise, directly fusing them with image features may introduce unreliable geometric information and degrade the original image representations. Based on the consistency and complementarity between the depth prior and image features, DPG-CD adopts a gated strategy for hierarchical fusion.
Specifically, a gated auxiliary injection module is proposed to compute the degree to which the depth prior features are injected into the original image features. While preserving image features, the method dynamically learns weighting coefficients to control the strength of depth prior injection, enabling adaptive fusion of reliable geometric cues from the depth prior features.

The image features and depth prior features obtained from the encoder are fused at each corresponding level. Specifically, for each level, the input image feature 
$\mathbf{F}_{i}^{img}$ and the depth prior feature
$\mathbf{F}_{i}^{edm}$ are first transformed through convolution operations.
\begin{equation}\label{eq:trans}
\tilde{\mathbf{F}}_{i}^{img}=
\operatorname{Conv}\!\left(\mathbf{F}_{i}^{img}\right), \tilde{\mathbf{F}}_{i}^{edm}=
\operatorname{Conv}\!\left(\mathbf{F}_{i}^{edm}\right).
 \end{equation}
where, $\operatorname{Conv}$ represent individual convolution layer.

The transformed multi-modal features are then concatenated along the feature dimension, and a spatial weighting map is generated through the gating module.
\begin{equation}\label{eq:gate}
\mathbf{M}_{i}=
\sigma\bigl(
\phi_{m}\bigl(
\operatorname{Concat}(\tilde{\mathbf{F}}_{i}^{edm}, \tilde{\mathbf{F}}_{i}^{img})
\bigr)
\bigr).
\end{equation}
where, $\phi_m$ denotes the gated mapping composed of convolution, normalization, and activation functions.  And $\sigma$ represents the Sigmoid function. 
$\mathbf{M_{i}} \in \mathbb{R}^{1 \times H_{i} \times W_{i}}$ denotes the injection weights of geometric features at different spatial locations, which are normalized to the range $[0, 1]$ through the activation function.

Then, based on the gating weights, the transformed image features and depth prior features are fused. 
\begin{equation}\label{eq:imgfuse}
\mathbf{F}_{i}^{fused}= \tilde{\mathbf{F}}_{i}^{img} + \mathbf{M_{i}} \odot \tilde{\mathbf{F}}_{i}^{edm}.
 \end{equation}
where, $\odot$ denotes element-wise multiplication. 

Finally, through a residual connection, the geometry-enhanced features are injected into the original image features. The resulting feature $\hat{\mathbf{F}}_{i}^{img}$ is then used as the image modality input for subsequent change feature extraction.
 \begin{equation}\label{eq:imgfinal}
\hat{\mathbf{F}}_{i}^{img}= \phi_r\!(\mathbf{F}_{i}^{fused}) +  \mathbf{F}_{i}^{img}.
 \end{equation}
where, $\phi_r$ denotes a feature refinement operation composed of convolution, batch normalization, and ReLU activation.

\subsection{Multi-Stage Cross-Temporal Cross-Modal Feature Fusion}
After obtaining hierarchical DSM features $\mathbf{F}^{dsm}$ from the encoder and the fused image features $\hat{\mathbf{F}}^{img}$ derived in Section \ref{sec:Gated}, DPG-CD proposes a multi-stage change feature extraction framework to enable the fusion of cross-modal features across bi-temporal phases. For different levels, as illustrated in Fig.~\ref{fig:framework}, a Convolutional Channel Attention Block (CCAB) and a Hierarchical Change Feature Extraction Block (HCFEB) are respectively employed.

\subsubsection{Convolutional Channel Attention Block (CCAB) }
To preserve fine-grained spatial details and local texture information, a CCAB is introduced for the first-level features to extract change-relevant representations.
The CCAB models local spatial interactions between the two epochs and employs a channel attention mechanism to adaptively select change-sensitive channels. By globally responding to these features, the block effectively suppresses redundant information in unchanged regions and enhances the separability between changed and background areas.

To retain the complete semantic representation of both epochs, the features from the two epochs are first concatenated along the channel dimension as the input to the block.
\begin{equation}\label{eq:ccabconcat}
\mathbf{X}= \mathrm{Concat}(\mathbf{F}_i^{dsm}, \hat{\mathbf{F}}_i^{img}) \in \mathbb{R}^{2C_i \times H_i\times W_i}.
 \end{equation}
 
Based on the concatenated bi-temporal features, convolutional enhancement is first performed.
\begin{equation}\label{eq:ccabenhance}
\mathbf{\tilde{X}}=\operatorname{GELU}(\operatorname{BN}(\operatorname{Conv}(\mathbf{X}))).
 \end{equation}
where $\text{GELU}$ and $\text{BN}$ are the activation and normalization layers.
 
To adaptively emphasize channel responses for change discrimination, this module introduces a lightweight channel attention mechanism. Specifically, global average pooling is first applied to compress the spatial dimensions into a global channel descriptor, and then a bottleneck mapping composed of two convolution layers is used to model inter-channel dependencies.
 \begin{equation}\label{eq:aatention}
\mathbf{a} = W_2\, \mathrm{GELU}\!(W_1\, \mathrm{GAP}(\mathbf{\tilde{X}})).
 \end{equation}
where, $\mathrm{GAP}$ denotes global average pooling, $W_1$ and  $W_2$ denote the dimensionality reduction and expansion transformations, respectively.

Subsequently, normalized channel weights are obtained through the Sigmoid function. Finally, these weights are applied to the feature map, and the output is obtained via a residual connection.
  \begin{equation}\label{eq:ccabfinal}
\mathbf{X}_{out}^{CCAB}  = \mathbf{X}+  \mathbf{\tilde{X}}\odot \mathbf{\mathbf{a}}.
 \end{equation}
 
\subsubsection{Hierarchical Change Feature Extraction Block}
\begin{figure*}[h]
\centering
\includegraphics[width=0.8\linewidth]{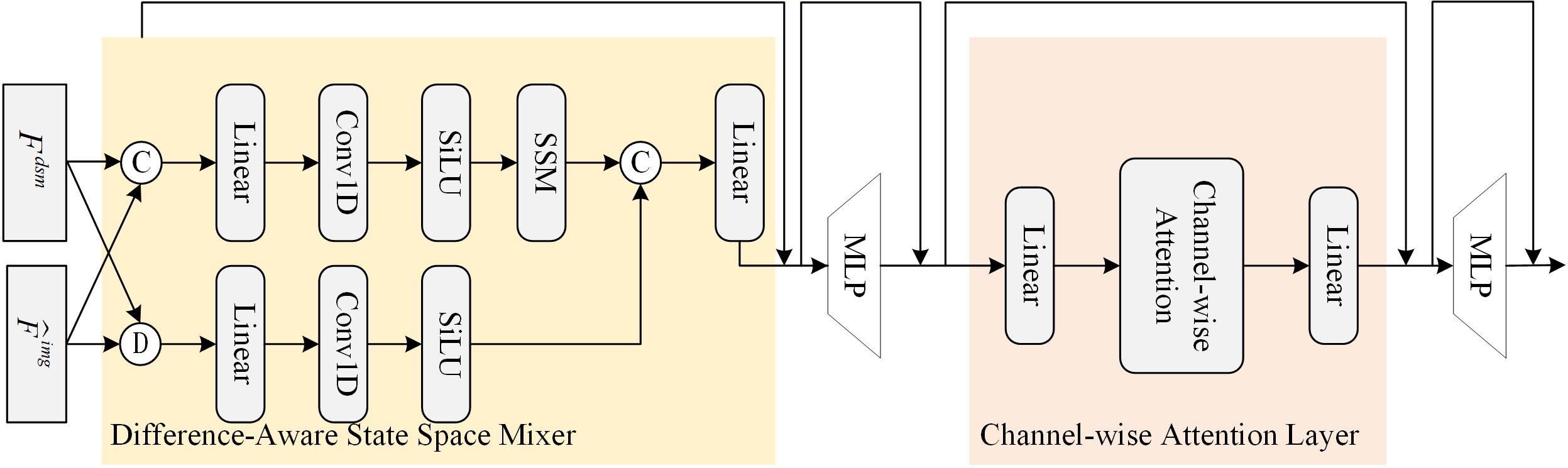}
\caption{Structure of the proposed hierarchical change feature extraction block.}
\label{fig:HCFE}
\end{figure*}
This paper adopts a hierarchical progressive change feature extraction module composed of two feature extraction layers to capture high-level change-related representations. The module includes a Difference-Aware State Space Mixer (DSSM) layer and a Cross-Channel Attention (CCA) layer, each followed by an MLP-based feature transformation.
The overall structure is illustrated in Fig. \ref{fig:HCFE}.

\textbf{Difference-Aware State Space Mixer(DSSM).}
To more effectively model the relationships and differences between bi-temporal features, this method proposes a difference-aware state space mixer module. This module adopts a dual-branch structure to capture global change features and enhance difference representations.

In the main branch, a state space selective scanning mechanism is employed to extract global change features, taking the concatenated bi-temporal features as input. The features are first reshaped into a sequential representation.
\begin{equation}\label{eq:dssminput}
\mathbf{X}^{seq}= \mathrm{Concat}(\mathbf{F}_i^{dsm}, \hat{\mathbf{F}}_i^{img}) \in \mathbb{R}^{L_i\times2C_i}.
 \end{equation}
 
The sequential features are first projected into a latent space, followed by a 1D convolution and a nonlinear activation to capture local contextual information.
\begin{equation}\label{eq:dssml}
\mathbf{S}=\mathrm{SiLU}(\mathrm{Conv}(\mathrm{Linear}(\mathbf{X}^{seq}))). 
 \end{equation}
where $\text{SiLU}$ is the activation function and $\text{Linear}$ represents the linear layer.

The mapped bi-temporal sequential features are then used to generate the parameters of a state-space model, and selective scanning \citep{gu2024mamba} is performed to extract features.
\begin{equation}\label{eq:scan}
\tilde{\mathbf{S}}=\mathrm{SCAN}(\mathbf{S}).
 \end{equation}
where, $\mathrm{SCAN}$ denotes the scanning operation.

In addition to the main branch for State Space Model (SSM), DSSM introduces an explicit difference-aware branch to enhance the model’s sensitivity to change-related feature.
The difference features are constructed by subtracting the original input features and are further enhanced through another set of 1D convolutions to capture local differences.
\begin{equation}\label{eq:changedssm}
\mathbf{D}=\mathrm{SiLU}\big(\mathrm{Conv}(\mathrm{Linear}(\hat{\mathbf{F}}_i^{img}- \mathbf{F}_i^{dsm})\big).
 \end{equation}

Subsequently, the outputs of the main branch and the difference-aware branch are concatenated to obtain difference-enhanced features.
 \begin{equation}\label{eq:dssmfinal}
\mathbf{X}_{out}^{DSSM} = \mathrm{Linear}\left(\mathrm{Concat}(\tilde{\mathbf{S}}, \mathbf{D})\right).
 \end{equation}
 
The DSSM is able to model long-range dependencies while explicitly enhancing bi-temporal difference information.

\textbf{Cross-Channel Attention(CCA).}
The features processed by the DSSM are further refined to strengthen change-related representations with  a Cross-Channel Attention (CCA) module. 
%By computing attention weights along the channel dimension, the module captures the differential relationships between features across temporal instances.

Given the feature $\tilde{\mathbf{X}}_{out}^{DSSM}$
extracted by the DSSM and MLP layer, query, key, and value are first generated through linear projections.
 \begin{equation}\label{eq:qkv}
\mathbf{Q},\mathbf{K},\mathbf{V}=\mathrm{Linear}\big(\tilde{\mathbf{X}}_{out}^{DSSM}\big).
 \end{equation}

Subsequently, the attention computation is shifted from the token dimension to the channel dimension, where multi-head scaled dot-product attention is performed.
  \begin{equation}\label{eq:atten}
\mathbf{X}_{out}^{CCA} = \mathrm{Linear}\left(\mathrm{Softmax}\left(\frac{\mathbf{Q}^{\top}\mathbf{K}}{\sqrt{d}}\right)\mathbf{V}^{\top}\right).
 \end{equation}
where, $d$ denotes the dimensionality of each attention head.

\subsection{Multi-task Decoder}

\subsubsection{UPerNet-based Multi-scale Feature Fusion}
As shown in Fig.~\ref{fig:framework}, the hierarchical features from the two epochs are processed through a multi-stage cross-temporal cross-modal feature fusion architecture. The first-stage change features are obtained by the CCA Block, while the remaining three stages are produced by the HCFE Block, resulting in hierarchical change features $\left\{ \mathbf{F}_{1}^{ch}, \mathbf{F}_{2}^{ch}, \mathbf{F}_{3}^{ch}, \mathbf{F}_{4}^{ch} \right\}$.

Subsequently, this paper adopts the UPerNet \citep{Xiao_2018_ECCV} decoding framework for multi-scale change features fusion.
%UPerNet combines the Pyramid Pooling Module (PPM) and the Feature Pyramid Network (FPN) to fuse multi-level features and generate prediction features.
\begin{equation}\label{eq:uper}
\mathbf{F}_{\mathrm{ch}}=\mathcal{D}_{\mathrm{UPer}}\!(\left\{ \mathbf{F}_{1}^{ch}, \mathbf{F}_{2}^{ch}, \mathbf{F}_{3}^{ch}, \mathbf{F}_{4}^{ch} \right\}).
\end{equation}
where $\mathcal{D}_{\mathrm{UPer}}$ denotes the UPerNet-based decoding architecture.

Finally, upsampling is applied to obtain the final change feature $\mathbf{F}_{\mathrm{ch}}$, which are then used to predict 2D semantic change and 3D height change, respectively.
\begin{equation}\label{eq:2d}
\hat{\mathbf{Y}}^{2\mathrm{d}}=\mathcal{H}_{2\mathrm{d}}(\operatorname{Up_{2d}}(\mathbf{F}_{\mathrm{ch}})).
\end{equation}
\begin{equation}\label{eq:3d}
\hat{\mathbf{Y}}^{3\mathrm{d}}=\mathcal{H}_{3\mathrm{d}}(\operatorname{Up_{3d}}(\mathbf{F}_{\mathrm{ch}})).
\end{equation}
where, $\operatorname{Up_{2d}}$ and $\operatorname{Up_{3d}}$ denote the upsampling layers, while $\mathcal{H}_{2\mathrm{d}}$ and $\mathcal{H}_{3\mathrm{d}}$ represent the 2D semantic change prediction head and the 3D height change prediction head, respectively.

\subsubsection{Auxiliary DSM Prediction Task}
In addition to obtaining 2D and 3D change features through bi-temporal feature fusion, this paper uses the DSM values corresponding to the T2 epoch as an auxiliary supervision and adopts a lightweight Feature Pyramid Network (FPN)-style decoder. The decoder takes the fused image features $\hat{\mathbf{F}}^{img}$ as input and performs channel alignment, top-down feature fusion, and multi-scale concatenation to integrate multi-level features and generate prediction features.
\begin{equation}\label{eq:light}
\mathbf{F}_{\mathrm{dsm}}=\mathcal{D}_{\mathrm{LP}}\!\left( \left\{ \hat{\mathbf{F}}_{1}^{img}, \hat{\mathbf{F}}_{2}^{img}, \hat{\mathbf{F}}_{3}^{img}, \hat{\mathbf{F}}_{4}^{img} \right\}\right).
\end{equation}
where $\mathcal{D}_{\mathrm{LP}}$ denotes the lightweight FPN-style decoder architecture.

The features obtained from the lightweight decoder are upsampled to predict the DSM values.
\begin{equation}\label{eq:dsmloss}
\hat{\mathbf{Y}}^{\mathrm{dsm}}=\mathcal{H}_{\mathrm{dsm}}\left(\operatorname{Up}_{dsm}\left(\mathbf{F}_{\mathrm{dsm}}\right)\right).
\end{equation}
where $\operatorname{Up_{dsm}}$ denotes the upsampling layer, while $\mathcal{H}_{dsm}$ represent the DSM values prediction head.

\subsection{Loss Function}
In the proposed cross-modal 2D and 3D change detection task, DPG-CD separately predicts 2D semantic changes and 3D height changes. In addition, DPG-CD predicts the DSM corresponding to the T2 epoch to constrain height regression in change regions and reduce fluctuations in height prediction for unchanged areas.

To avoid additional data acquisition costs, the ground-truth supervision for DSM is derived from the T1 DSM and the 3D height change ground truth.
The final loss function consists of the 2D semantic change loss, the 3D height change loss, and the DSM prediction loss.

The 2D semantic change loss is formulated as a weighted cross-entropy loss $\mathcal{L}_{\mathrm{wCE}}$.
\begin{equation}\label{eq:wce}
\begin{aligned}
\mathcal{L}_{\mathrm{wCE}}
&=
-\frac{1}{N}\sum_{n=1}^{N}
w_{l_n}\,\log\!\left(
\frac{\exp(z_{n,l_n})}{\sum_{k=1}^{C}\exp(z_{n,k})}
\right), \\
&\qquad l_n \in \{0,1,\dots,C-1\}.
\end{aligned}
\end{equation}
where, $N$ denotes the total number of pixels, $l_n$ denotes the integer class label of the 
$n$-th pixel, $z_{n,k}$ represents the logit for class $k$, and 
$w_{l_n}$  is the weight corresponding to the ground-truth class.

To preserve accurate building structures and boundaries, the 3D height loss is composed of an MSE loss $\mathcal{L}_{\mathrm{MSE}}$ and a gradient loss $\mathcal{L}_{\mathrm{grad}}$. The 3D MSE is computed over the entire image, while the gradient loss is applied only within change regions.

\begin{equation}\label{eq:mse}
\mathcal{L}_{\mathrm{MSE}}
=
\frac{1}{N}\sum_{n=1}^{N}\left\| \hat{y}_n - y_n \right\|^{2}.
\end{equation}

where, $y_n$ is the ground truth of the 
 $n$-th pixel, and $\hat{y}_n$  is the corresponding prediction.
\begin{equation}\label{eq:gradloss}
\begin{split}
\mathcal{L}_{\mathrm{grad}}
&=
\frac{1}{N_{ch}}\sum_{i,j}
\Bigl(
\bigl|
(D_{i,j+1}-D_{i,j})-(T_{i,j+1}-T_{i,j})
\bigr| \\
&\qquad\qquad+
\bigl|
(D_{i+1,j}-D_{i,j})-(T_{i+1,j}-T_{i,j})
\bigr|
\Bigr)
\end{split}.
\end{equation}

where, $D$ denotes the predicted map, $T$ represents the ground truth map, and 
$N_{ch}$ is the total number of pixels in the change regions.

The auxiliary DSM supervision is defined using an MSE loss $\mathcal{L}_{\mathrm{MSE}}$.

The final overall loss is composed of a weighted cross-entropy loss $\mathcal{L}_{\mathrm{wCE}}$ for 2D semantic change, an MSE loss $\mathcal{L}_{\mathrm{mse}}^{3d}$ for 3D height change, a gradient loss $\mathcal{L}_{\mathrm{grad}}$ for 3D height changes in change regions, and an MSE loss $\mathcal{L}_{\mathrm{mse}}^{dsm}$ for DSM prediction .
\begin{equation}\label{eq:allloss}
\mathcal{L}_{\mathrm{total}}=\lambda_{\mathrm{wCE}} \cdot \mathcal{L}_{\mathrm{wCE}} + \lambda_{\mathrm{mse}}^{3d} \cdot \mathcal{L}_{\mathrm{mse}}^{3d} + \cdot \lambda_{\mathrm{grad}} \cdot \mathcal{L}_{\mathrm{grad}} +\lambda_{\mathrm{mse}}^{dsm} \cdot \mathcal{L}_{\mathrm{mse}}^{dsm}.
\end{equation}
where, $\lambda_{\mathrm{wCE}}$, $ \lambda_{\mathrm{mse}}^{3d}$, $\lambda_{\mathrm{grad}}$, and $\lambda_{\mathrm{mse}}^{dsm}$ denote the weights of the different loss terms.
\section{Experimental Settings \label{sec_exp_settings}}
\subsection{Metrics}
For the 2D semantic change detection task, this paper evaluates performance using per-class Intersection over Union (IoU), mean IoU (mIoU), and mean F1 score (mF1). Specifically, the IoU and F1 score are first computed for each semantic class and then averaged over all classes.
\begin{equation}\label{eq:iou}
\mathrm{IoU}_c = \frac{\mathrm{TP}_c}{\mathrm{TP}_c +\mathrm{FP}_c + \mathrm{FN}_c}.
\end{equation}
\begin{equation}\label{eq:f1}
\mathrm{F1}_c = \frac{2\mathrm{TP}_c}{2\mathrm{TP}_c + \mathrm{FP}_c + \mathrm{FN}_c}.
\end{equation}
where $\mathrm{TP}_c$, $\mathrm{FP}_c$, and $\mathrm{FN}_c$ denote the numbers of true positive, false positive, and false negative pixels for class $c$, respectively.

To evaluate the performance of 3D height changes, this paper adopts four metrics, including  Mean Absolute Error (MAE), Root Mean Square Error (RMSE), change-region RMSE (cRMSE), and change-region relative error (cRel).
Specifically, RMSE and MAE measure the overall numerical differences between the predicted height change map and the ground truth over the entire image, while cRMSE and cRel are computed within change regions to assess absolute and relative errors, respectively.
\begin{equation}\label{eq:mae}
\mathrm{MAE} = \frac{1}{N}\sum_{(i,j)\in \Omega}\left|\hat{H}_{ij}-H_{ij}\right|.
\end{equation}
\begin{equation}\label{eq:rmse}
\mathrm{RMSE} = \sqrt{\frac{1}{N}\sum_{(i,j)\in \Omega}\left(\hat{H}_{ij}-H_{ij}\right)^2}.
\end{equation}

\begin{equation}\label{eq:crmse}
\mathrm{cRMSE} = \sqrt{\frac{1}{N_c}\sum_{(i,j)\in \Omega_c}\left(\hat{H}_{ij}-H_{ij}\right)^2}.
\end{equation}
\begin{equation}\label{eq:rel}
\mathrm{cRel} = \frac{1}{N_c}\sum_{(i,j)\in \Omega_c} \frac{\left|\hat{H}{ij}-H{ij}\right|}{|H_{ij}|}
\end{equation}
where, $\hat{H}_{ij}$ and $H_{ij}$
denote the predicted and ground-truth height change values at pixel $(i,j)$, respectively. $\Omega$ and 
$\Omega_c$ represent the sets of all pixels and change-region pixels.

\subsection{Experimental datasets}
\begin{figure}[]
\centering
\includegraphics[width=\linewidth]{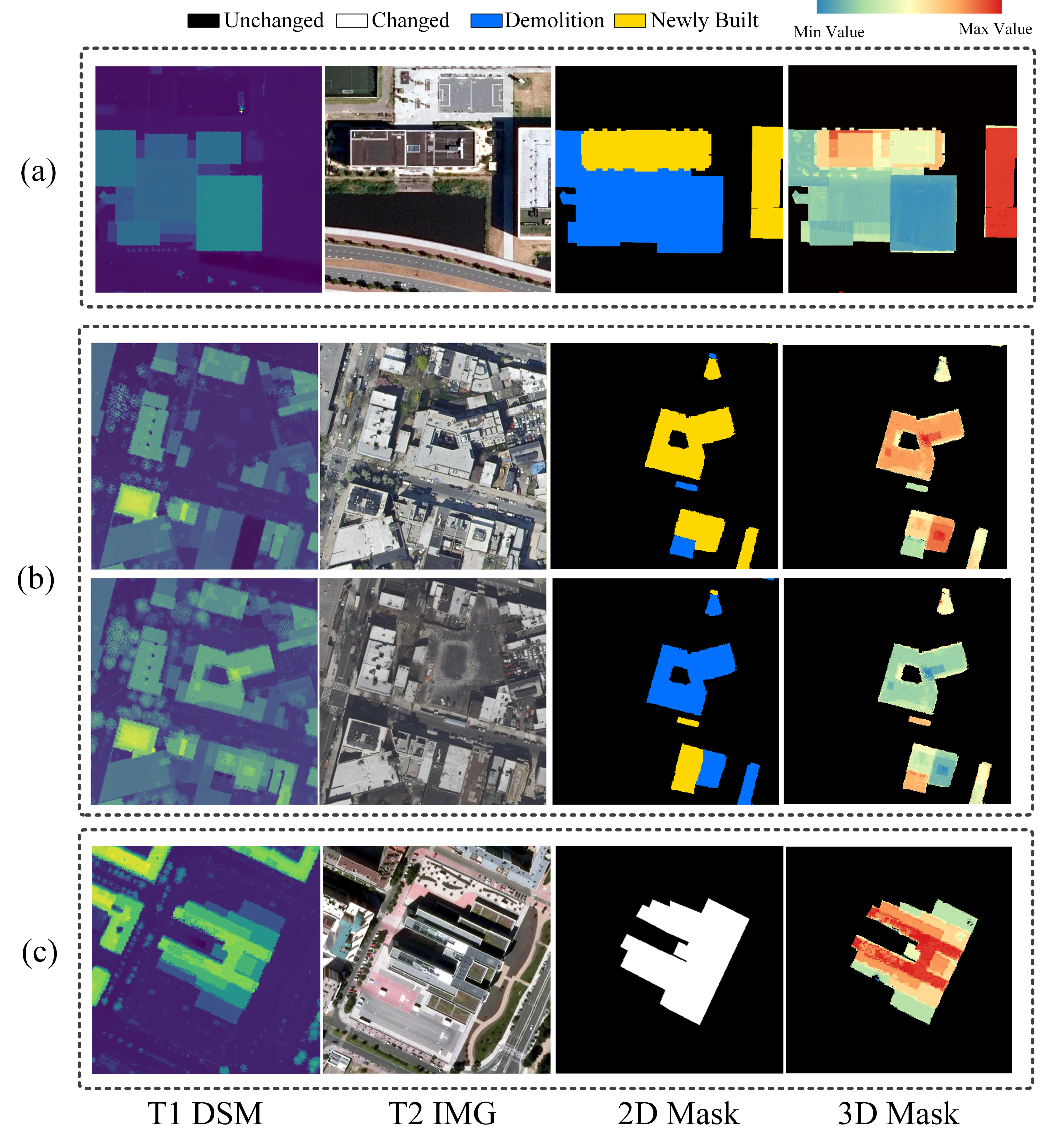}
\caption{Examples of inputs and labels from the multi-modal change detection dataset: (a)3DCD dataset; (b)Hi-BCD dataset; (c)NYC-MMCD dataset.}
\label{fig:dataset}
\end{figure}

To evaluate the effectiveness of the proposed cross-modal 2D and 3D change detection method, we conducted experiments on three real-world cross-modal datasets, including two publicly available datasets and the newly proposed cross-modal dataset in this paper.
The statistics and sample visualizations of the three datasets are presented in Tab. \ref{tab:datasets} and Fig. \ref{fig:dataset}, respectively.

\begin{table}[]
\centering
\caption{Statistics of existing multi-modal multi-task change detection datasets.}
\label{tab:datasets}
\resizebox{\columnwidth}{!}{%
\begin{tabular}{l|ccccc}
\hline
Name     & Image pairs &\begin{tabular}[c]{@{}c@{}}Tile size \\ change type  (pixels) \end{tabular}  & Resolution & \begin{tabular}[c]{@{}c@{}}Classes of \\ change type\end{tabular} & \begin{tabular}[c]{@{}c@{}}Percentage of\\   change pixels (\%)\end{tabular} \\ \hline
Hi-BCD   & 1500        & 1000 × 1000 & 0.25 m     & 2  & 1.14  
\\
NYC-MMCD & 1000        & 500× 500    & 0.3 m      & 2   & 2.46                                                   \\ 
3DCD     & 472         & 400 × 400   & 0.5/1 m    & 1    & 4.67  
\\ \hline
\end{tabular}%
}
\end{table}

\subsubsection{Hi-BCD dataset}
\citet{liu2024transformer} introduced the Hi-BCD cross-modal dataset for building change detection, which jointly identifies 2D semantic and 3D height changes using pre-event DSM and post-event aerial imagery. The dataset covers three Dutch cities: Amsterdam, Rotterdam, and Utrecht. Specifically, Hi-BCD comprises a total of 1500 DSM–image pairs, equally distributed with 500 samples per city. Each sample has a size of 1000 × 1000 and a spatial resolution of 0.25 m.
The dataset categorizes two types of building changes into newly-built and demolition, providing both semantic change labels and their corresponding height change values.

\subsubsection{NYC-MMCD dataset}
\citet{zhang2026me} presented a semantic change detection dataset based on bi-temporal point clouds, wherein the post-event point cloud is annotated with classes including unchanged, newly built,demolition, and new clutters. In this work, aerial imagery is further integrated, and the point cloud data were converted into DSM representations with a spatial resolution of 0.3 m.
The original point cloud data were collected in 2014 and 2017, while the aerial imagery was sourced from the New York Statewide Digital Orthoimagery Program. In this study, imagery from 2014 and 2018 was employed, with its spatial resolution resampled to match that of the DSM at 0.3 m.

As depicted in Fig. \ref{fig:dataset} (b), two sets of cross-modal asymmetric bitemporal pairs are established: (2014 DSM, 2018 imagery) and (2018 DSM, 2014 imagery). Corresponding 2D semantic changes and 3D height changes are annotated for the image modality, where the 2D semantic change categories include unchanged, newly built, and demolition.
Due to the diverse scenarios in the NYC-MMCD dataset, joint 2D-3D change detection becomes more challenging. For example, densely built areas lead to issues such as shadows and occlusions in the image modality.

\subsubsection{3DCD dataset}
3DCD dataset \citep{marsocci2023inferring} is a publicly available dataset designed for joint 2D and 3D change detection utilizing bi-temporal optical imagery. It encompasses diverse landscapes within Valladolid, Spain, ranging from the historic center to urban built-up areas and surrounding commercial districts. The 2D semantic labels are defined as binary change labels.

The dataset comprises 472 samples, each of which encompasses bi-temporal aerial orthophotos, DSMs, and their corresponding 2D semantic and 3D height change ground truths. The orthophotos were acquired in 2010 and 2017, with a size of 400 × 400 pixels at a spatial resolution of 0.5 m. And the DSMs have a size of 200 × 200 pixels with a resolution of 1.0 m. 
%The dataset also provides semantic change labels and height change information.
To evaluate the performance of the proposed cross-modal change detection method, we use the 2010 DSM and the 2017 aerial orthophotos from this dataset to construct cross-modal inputs.
\subsection{Compared methods}
To evaluate the performance of DPG-CD, eight state-of-the-art (SOTA) change detection methods are selected for comprehensive comparison. These methods are categorized into two groups. The first group comprises methods specifically designed for joint predict 2D and 3D changes: MTBIT \citep{marsocci2023inferring}, MMCD \citep{liu2024transformer}, and HATFormer \citep{liu2025hatformer}. Notably, MMCD and HATFormer utilize cross-modal fusion, whereas MTBIT performs multi-dimensional change analysis based on bi-temporal optical imagery. 
Additionally, several representative and advanced 2D change detection frameworks are selected and adapted to support simultaneous 3D height change estimation, including FC-Siam-conc \citep{daudt2018fully}, SNUNet \citep{fang2021snunet}, ChangeFormer \citep{bandara2022transformer}, AMTNet \citep{liu2023attention}, and ChangeMamba \citep{chen2024changemamba}. 

\subsection{Implementation details}
Following the experimental protocols of the cross-modal change detection methods MMCD \citep{liu2024transformer} and HATFormer \citep{liu2025hatformer}, all methods are pre-trained on the LEVIR-CD dataset \citep{chen2020spatial}. 
In the experiments, this paper leverages the MambaVision-T architecture initialised with pre-trained weights \citep{Hatamizadeh_2025_CVPR}. In Eq. \ref{eq:allloss}, the weights $\lambda_{\mathrm{wCE}}$, $ \lambda_{\mathrm{mse}}^{3d}$, $\lambda_{\mathrm{grad}}$, and $\lambda_{\mathrm{mse}}^{dsm}$ are set to $[1,1,0.2,1]$. To address class imbalance, the weights for the weighted cross-entropy (wCE) loss were assigned as 0.95 for the changed class and 0.05 for the unchanged class.
The Hi-BCD and NYC-MMCD datasets are cropped into patches of size 512 × 512 pixels, while the input size for the 3DCD dataset is 400 × 400 pixels. These models are trained for 200, 250, and 300 epochs on the Hi-BCD, NYC-MMCD, and 3DCD datasets, respectively.

\section{Experimental results \label{sec_exp_re}}
\subsection{Results on Hi-BCD dataset}
\begin{figure*}[h]
\centering
\includegraphics[width=\linewidth]{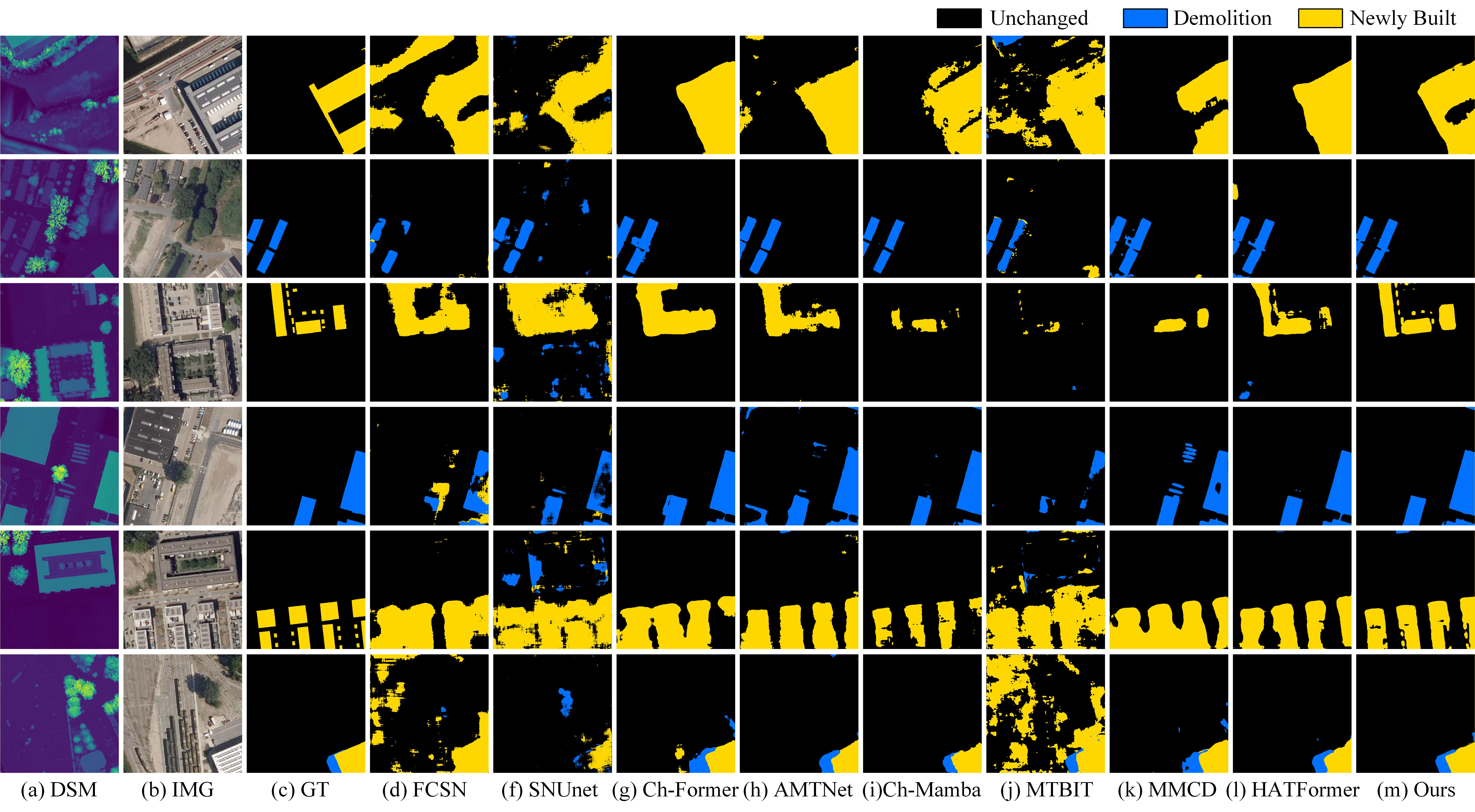}
\caption{Comparison of 2D semantic change detection results of all methods on Hi-BCD dataset.}
\label{fig:hibcd_2d}
\end{figure*}
\begin{figure*}[h]
\centering
\includegraphics[width=\linewidth]{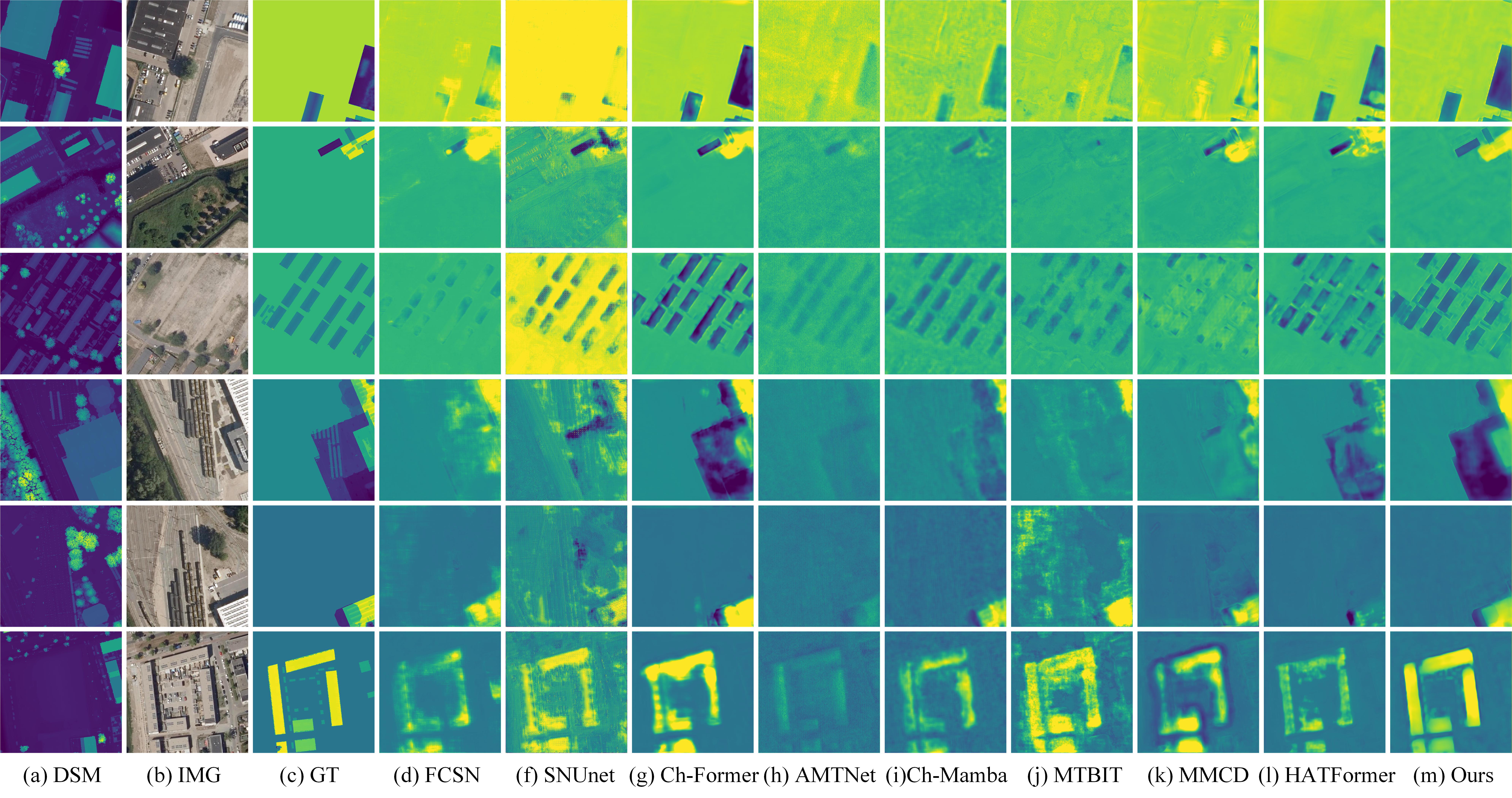}
\caption{Comparison of 3D height change detection results of all methods on the Hi-BCD dataset.}
\label{fig:hibcd_3d}
\end{figure*}
\begin{figure}[]
\centering
\includegraphics[width=\linewidth]{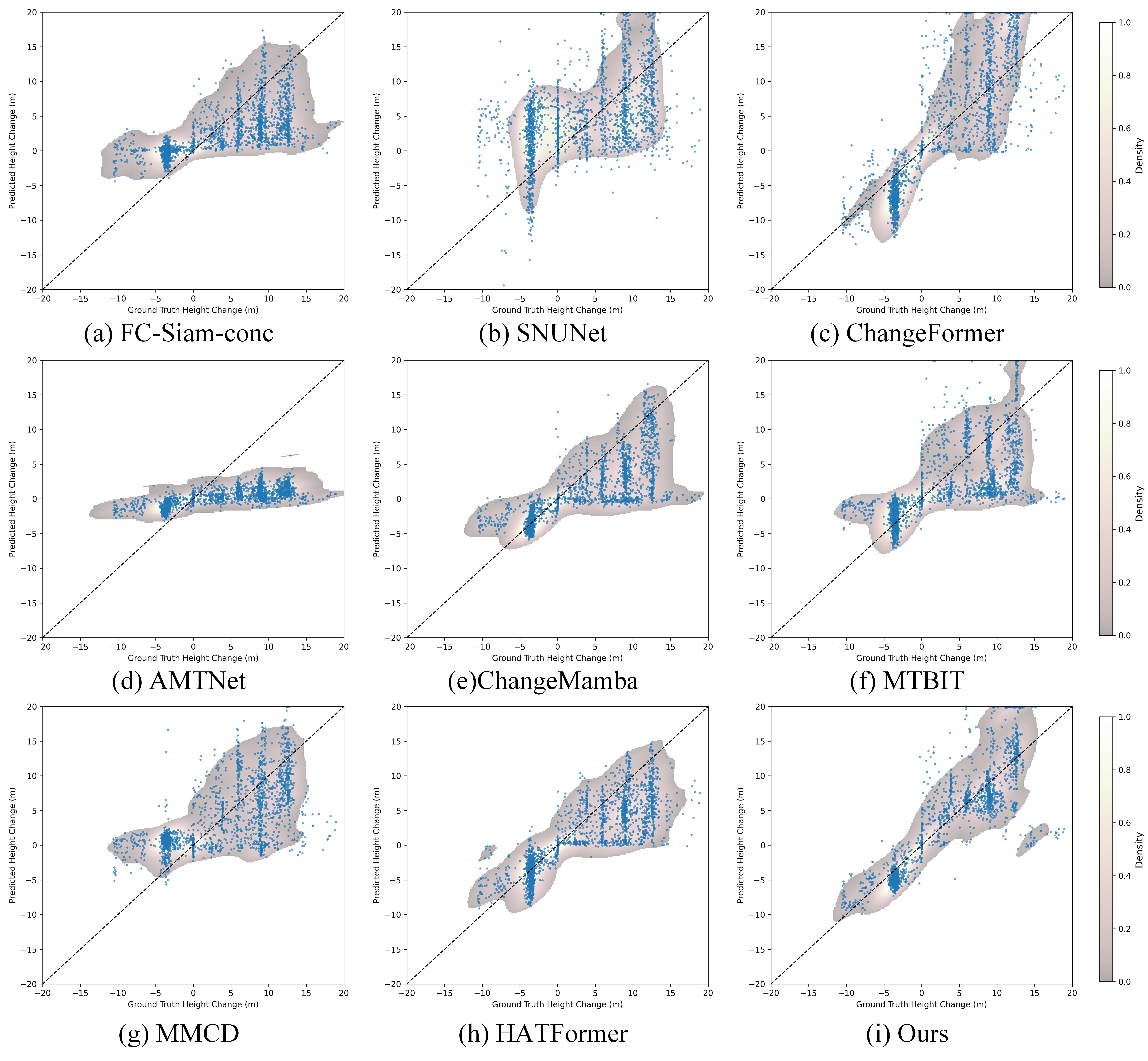}
\caption{Scatter plots with KDE visualization of the relationship between predicted and ground-truth values for all methods on the Hi-BCD dataset.}
\label{fig:height-hibcd}
\end{figure}

\begin{table*}[]
\centering
\caption{Quantitative evaluation results on Hi-BCD dataset.}
\label{tab:hibcd}
\begin{tabular}{l|lllll|llll}
\hline
\multicolumn{1}{c|}{\multirow{2}{*}{Method}}                & \multicolumn{5}{c|}{2D task metric (\%)}      & \multicolumn{4}{c}{3D task metric (m)} \\ \cline{2-10} 
\multicolumn{1}{c|}{}                                       & IoU\_un$\uparrow$ & IoU\_d$\uparrow$ & IoU\_n$\uparrow$ & miou\_ch$\uparrow$ & mF1 $\uparrow$   & MAE$\downarrow$     & RMSE$\downarrow$      & cRMSE$\downarrow$     & cRel$\downarrow$      \\ \hline
FC-Siam-conc\citep{daudt2018fully}         & 96.73   & 15.31  & 18.25  & 16.78    & 28.71  & 0.317    & 1.435   & 8.524   & 1.707   \\
SNUNet\citep{fang2021snunet}               & 97.63   & 15.98  & 22.04  & 19.01    & 54.16  & 1.418    & 2.314   & 8.805   & 4.814   \\
ChangeFormer\citep{bandara2022transformer}  & 98.79	&55.68	&36.62	&46.15	&62.57	&0.209	&1.357	&8.266	&1.417       \\
AMTNet\citep{liu2023attention}   & 98.91   & 45.75  & 33.03  & 39.39    & 70.63  & 0.432    & 1.564   & 9.315   & 1.565   \\
ChangeMamba\citep{chen2024changemamba}     & 99.06   & 51.81  & 25.43  & 38.62    & 69.44  & 0.421    & 1.462   & 3.317   & 1.668   \\
MTBIT\citep{marsocci2023inferring}         & 98.25   & 29.94  & 21.34  & 25.64    & 60.13  & 0.446    & 1.561   & 8.932   & 1.610    \\
MMCD\citep{liu2024transformer}             & 98.62    & 43.23   & 26.82   & 35.03   & 67.33 & 0.367    & 1.492   & 8.815   & 1.689   \\
HATFormer\citep{liu2025hatformer}          & 98.89   & 51.93  & 33.04  & 42.48    & 59.01  & 0.277    & 1.399   & 8.704   & \textbf{1.260}    \\
Ours                                                        & \textbf{99.34}   & \textbf{64.10}   & \textbf{54.64}  & \textbf{59.37}    & \textbf{82.82}  & \textbf{0.225}   & \textbf{1.104}  & \textbf{6.831}  & 1.425  \\ \hline
\end{tabular}
\end{table*}
Tab. \ref{tab:hibcd} summarizes the quantitative results for 2D semantic change and 3D height change on the Hi-BCD dataset.
The proposed DPG-CD framework demonstrates superior performance across both tasks. 
Fig. \ref{fig:hibcd_2d} and Fig. \ref{fig:hibcd_3d} provide qualitative comparision of 2D and 3D change results under diverse and challenging urban scenarios.

As visualized in Fig. \ref{fig:hibcd_2d}, the proposed DPG-CD framework demonstrates higher geometric accuracy and enhanced boundary completeness in semantic change segmentation. In contrast to comparative methods, which frequently produce ambiguous or fragmented boundaries, DPG-CD exhibits a robust capability in reconstructing fine-grained spatial details, effectively preserving the linear contours and orthogonal geometries of changed buildings.
Within high-density urban areas, many comparative methods suffer from omission errors and under-segmentation artifacts. Conversely, DPG-CD can precisely delineate independent buildings in close proximity and accurately capture subtle structural changes.
Furthermore, scattered noise pixels are observed in comparative methods in regions plagued by vegetation and shadow interference. 
DPG-CD effectively aligns DSM features with image features, enabling accurate change detection while suppressing modality-specific noise, thereby significantly reducing false positives.

Regarding the 3D height change prediction task, the proposed method achieves superior estimation accuracy, reflecting an effective synergy between the 2D semantic and 3D geometric tasks.
As visualized in Fig. \ref{fig:hibcd_3d}, comparison methods often produce noise in unchanged regions, leading to erroneous height change predictions. Conversely, DPG-CD exhibits significantly greater stability in color and texture within unchanged regions. This capability underscores the model's robustness in identifying unchange areas, ensuring high consistency between the predicted 3D height changes and the corresponding 2D semantic change masks.

Furthermore, the proposed DPG-CD framework demonstrates efficacy in preserving building morphology. For instance, the rooftop of an individual structure should display homogeneous and continuous elevation profiles. As shown in Fig. \ref{fig:hibcd_3d}, the visualized height prediction maps generated by our method exhibit a color trend closer to the ground truth distribution, indicating that the regression values are more aligned with the actual height changes. This indicates that the method has a strong capability to capture overall building structures and enables pixel-level accurate elevation reconstruction.
As shown in the last row of Fig. \ref{fig:hibcd_3d}, the predicted height maps also maintain sharp building boundaries.
Moreover, even within dense urban clusters, small-scale architectural details and open spaces are accurately delineated.

To further evaluate the regressive fidelity, Fig. \ref{fig:height-hibcd} visualize the correlation between predicted and ground-truth values via Kernel Density Estimation (KDE)-augmented scatter plots, where the color intensity represents the local sample density. 
The results indicate that methods initially designed for joint 2D-3D change detection often yield better performance in height change prediction. This suggests that while 3D height changes is intrinsically linked to 2D semantic change detection, it necessitates specialized mechanisms to achieve high-precision numerical regression.
As evidenced by the scatter plots, the proposed DPG-CD framework achieves the highest fitting accuracy. The tighter clustering of data points around the identity line ($y=x$) demonstrates that DPG-CD framework effectively mitigates the cross-modal heterogeneity and noise, yielding more precise and stable height change predictions. In contrast, other methods exhibit significant dispersion with broader density distributions, whereas DPG-CD maintains a highly concentrated distribution toward the diagonal, which directly correlates with its lower cRMSE.

\subsection{Results on NYC-MMCD dataset}
\begin{figure*}[]
\centering
\includegraphics[width=\linewidth]{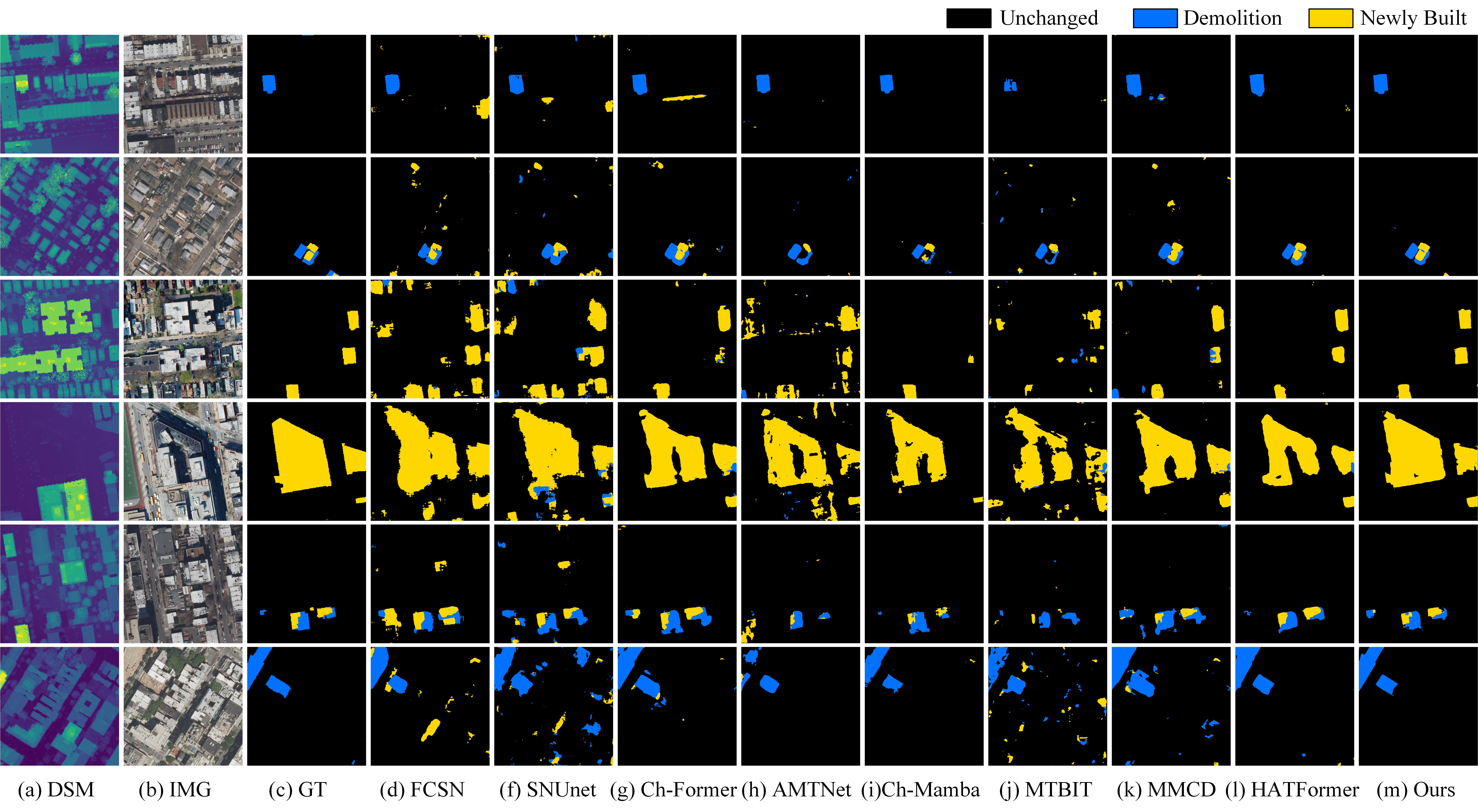}
\caption{Comparison of 2D semantic change detection results of all methods on NYC-MMCD dataset.}
\label{fig:nyc_2d}
\end{figure*}
\begin{figure*}[]
\centering
\includegraphics[width=\linewidth]{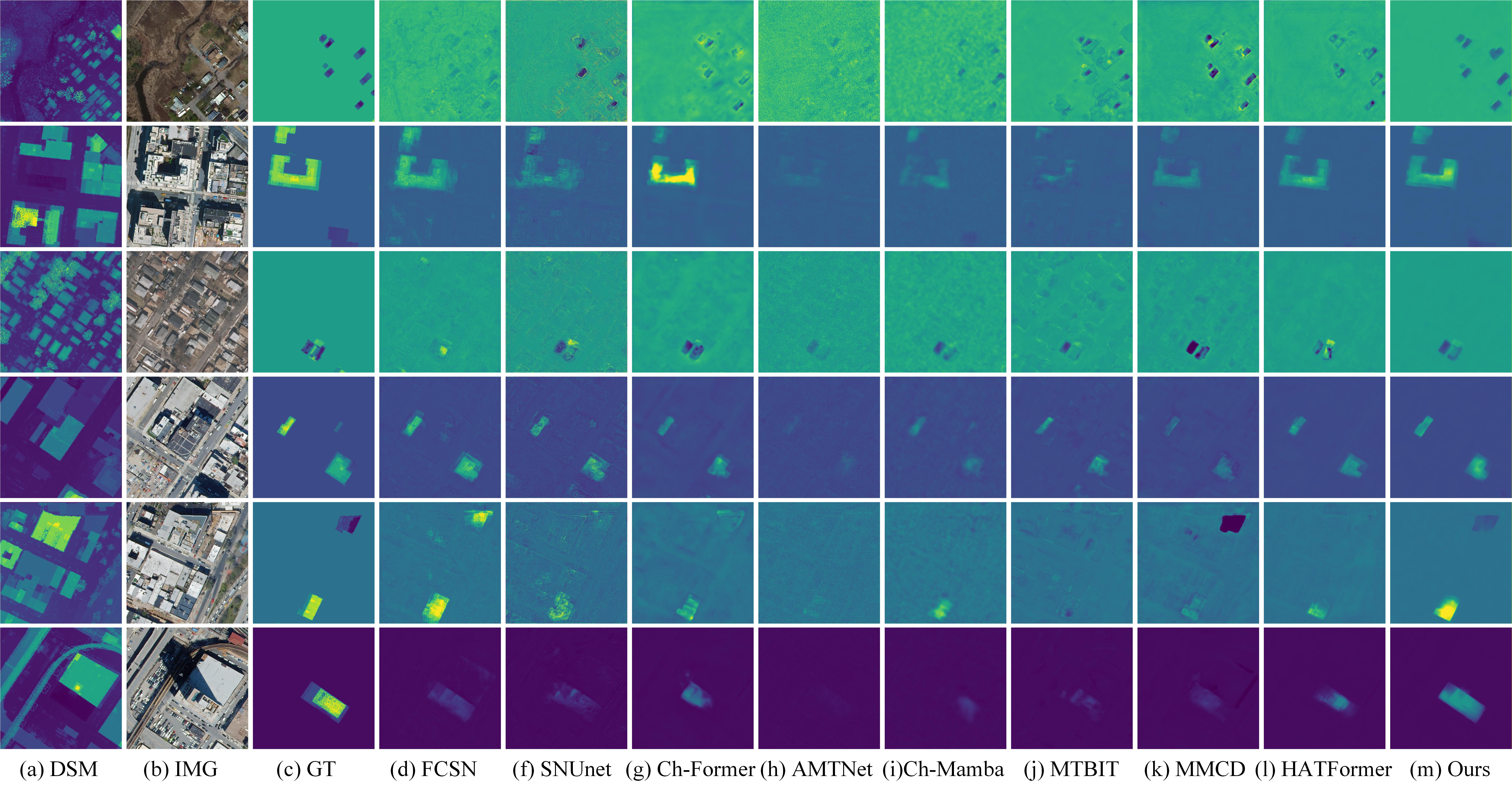}
\caption{Comparison of 3D height change detection results of all methods on NYC-MMCD dataset.}
\label{fig:nyc_3d}
\end{figure*}
\begin{figure}[]
\centering
\includegraphics[width=\linewidth]{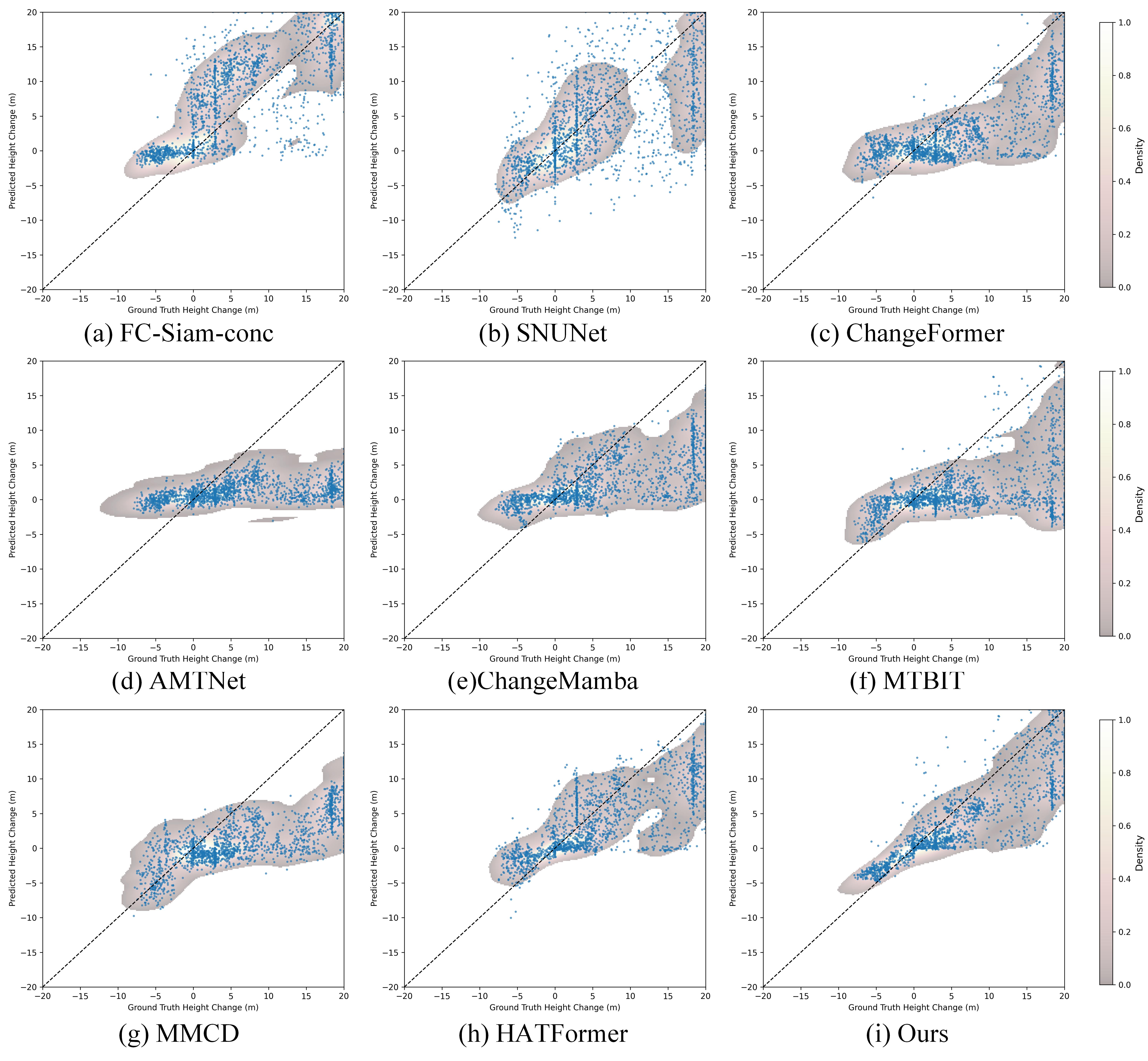}
\caption{Scatter plots with KDE visualization of the relationship between predicted and ground-truth values for all methods on the NYC-MMCD dataset.}
\label{fig:height-nyc}
\end{figure}

\begin{table*}[]
\centering
\caption{Quantitative evaluation results on NYC-MMCD dataset.}
\label{tab:NYC}
\begin{tabular}{l|ccccc|cccc}
\hline
\multicolumn{1}{c|}{\multirow{2}{*}{Method}}                & \multicolumn{5}{c|}{2D task metric (\%)}      & \multicolumn{4}{c}{3D task metric (m)} \\ \cline{2-10} 
\multicolumn{1}{c|}{}                          & IoU\_un$\uparrow$ & IoU\_n$\uparrow$ & IoU\_d$\uparrow$ & miou\_ch$\uparrow$ & mF1 $\uparrow$   & MAE  $\downarrow$   & RMSE$\downarrow$     & cRMSE$\downarrow$    & cRel$\downarrow$     \\ \hline
FC-Siam-conc\citep{daudt2018fully}         & 90.01   & 19.30   & 26.68  & 22.99    & 37.24  & 0.583    & 2.396   & 10.937  & 4.419   \\
SNUNet\citep{fang2021snunet}               & 95.08   & 18.48  & 22.49  & 20.48    & 55.13 & 1.457    & 2.914   & 11.07   & 5.418   \\
ChangeFormer\citep{bandara2022transformer} & 98.07   & 41.43  & 33.41  & 37.42    & 69.23 & 0.602    & 2.228   & 10.047  & 2.714   \\
AMTNet\citep{liu2023attention}             & 97.82   & 28.17  & 24.07  & 26.12    & 60.55  & 0.652    & 2.541   & 11.918  & 2.033   \\
ChangeMamba\citep{chen2024changemamba}     & 98.33   & 30.48  & 28.24  & 29.36    & 63.30   & 0.790     & 2.372   & 9.426   & 2.358   \\
MTBIT\citep{marsocci2023inferring}         & 91.72   & 23.12  & 19.72  & 21.42    & 35.25 & 0.575    & 2.498   & 11.838  & 1.991   \\
MMCD\citep{liu2024transformer}             & 92.14   & 36.16  & 29.41  & 32.78    & 49.28  & 0.550     & 2.218   & 10.512  & 2.640    \\
HATFormer\citep{liu2025hatformer}          & 92.69   & 39.53  & 32.22  & 35.88    & 52.70   & 0.393    & 2.141   & 10.108  & 2.204   \\
Ours & \textbf{98.43}   & \textbf{44.10}   & \textbf{33.61}  & \textbf{ 38.86}    &\textbf{ 70.25}  & \textbf{0.276}   & \textbf{1.908}  & \textbf{9.344}  & \textbf{1.576}  \\ \hline
\end{tabular}
\end{table*}
Tab. \ref{tab:NYC} summarizes the quantitative assessment of 2D semantic change and 3D height change prediction results of the proposed method and comparison methods on the NYC-MMCD dataset. Relative to the Hi-BCD dataset, the NYC-MMCD dataset is of moderate size yet involves more complex change scenarios.
Furthermore, the urban morphology in NYC-MMCD is predominantly composed of taller structures with significantly higher building density. Consequently, textures and shadow occlusions in the image modality introduce greater interference in non-change regions, making the change detection task more challenging.

Fig. \ref{fig:nyc_2d} and Fig. \ref{fig:nyc_3d} provide qualitative assessments of 2D and 3D results on NYC-MMCD dataset.
Regarding the 2D change task, the proposed DPG-CD framework exhibits efficacy in compact urban environments. Through multi-modal feature fusion, the model not only maintains the structural integrity of large buildings but also demonstrates robustness in fine-grained feature extraction and false-alarm suppression. Notably, DPG-CD outperforms competing approaches in the delineation of boundary details. For instance, the fourth row of Fig. \ref{fig:nyc_2d}, which features a newly constructed large-scale structure with a complex polygonal geometry, the segmentation prediction generated by DPG-CD achieves the highest degree of completeness.
Due to the small spacing between buildings and the influence of illumination and shadows, it is easy to misclassify adjacent unchanged regions as changed areas or confuse demolition with newly built categories.
Conversely, the second and fifth rows of Fig. \ref{fig:nyc_2d} illustrate highly complex downtown scenarios with extremely dense buildings, where both demolition and new construction occur simultaneously.  
%Conversely, as shown in the second row of Fig. \ref{fig:nyc_2d}, the proposed method successfully decouples adjacent changes belonging to distinct semantic classes, demonstrating robust semantic consistency. 
Furthermore, while many methods exhibit substantial spurious noise and false detections in regions plagued by complex shadows and non-building interference, DPG-CD manifests noise mitigation and error suppression.

In the 3D height change detection task, the proposed method achieves the lowest errors across all metrics. On the NYC-MMCD dataset, it effectively distinguishes invariant regions and generates uniform height estimates within the identified change areas.
As shown in the first row of Fig. \ref{fig:nyc_3d}, the proposed model can precisely capture subtle height changes of small-scale buildings. The predicted changes are not only spatially accurate but also exhibit height values consistent with the ground truth, demonstrating the model’s capability for high-fidelity local feature extraction.
Furthermore, as illustrated in the second row, the method produces consistent height predictions and preserves clear building contours, even for buildings with complex and irregular morphologies.

As depicted in the Fig. \ref{fig:height-nyc}, the scatter plots and KDE visualizations of height prediction results on the NYC-MMCD dataset indicate a more challenging scenario compared to the Hi-BCD dataset.
This is primarily attributed to the wider dynamic range and the more heterogeneous distribution of vertical structures in New York’s high-density urban environments.
Compared with comparison methods, many approaches exhibit noticeable bias when handling large height changes of tall buildings, making it difficult to accurately capture significant height changes.
In contrast, the proposed DPG-CD framework maintains stronger regression consistency on the NYC-MMCD dataset. In particular, it produces fewer residuals in unchanged regions around 0 m, demonstrating improved stability and accuracy in height prediction. The higher concentration of samples along the identity line underscores DPG-CD’s robustness in mitigating the complex cross-modal interference characteristic of large-scale urban landscapes.

\subsection{Results on 3DCD dataset}

\begin{figure*}[]
\centering
\includegraphics[width=\linewidth]{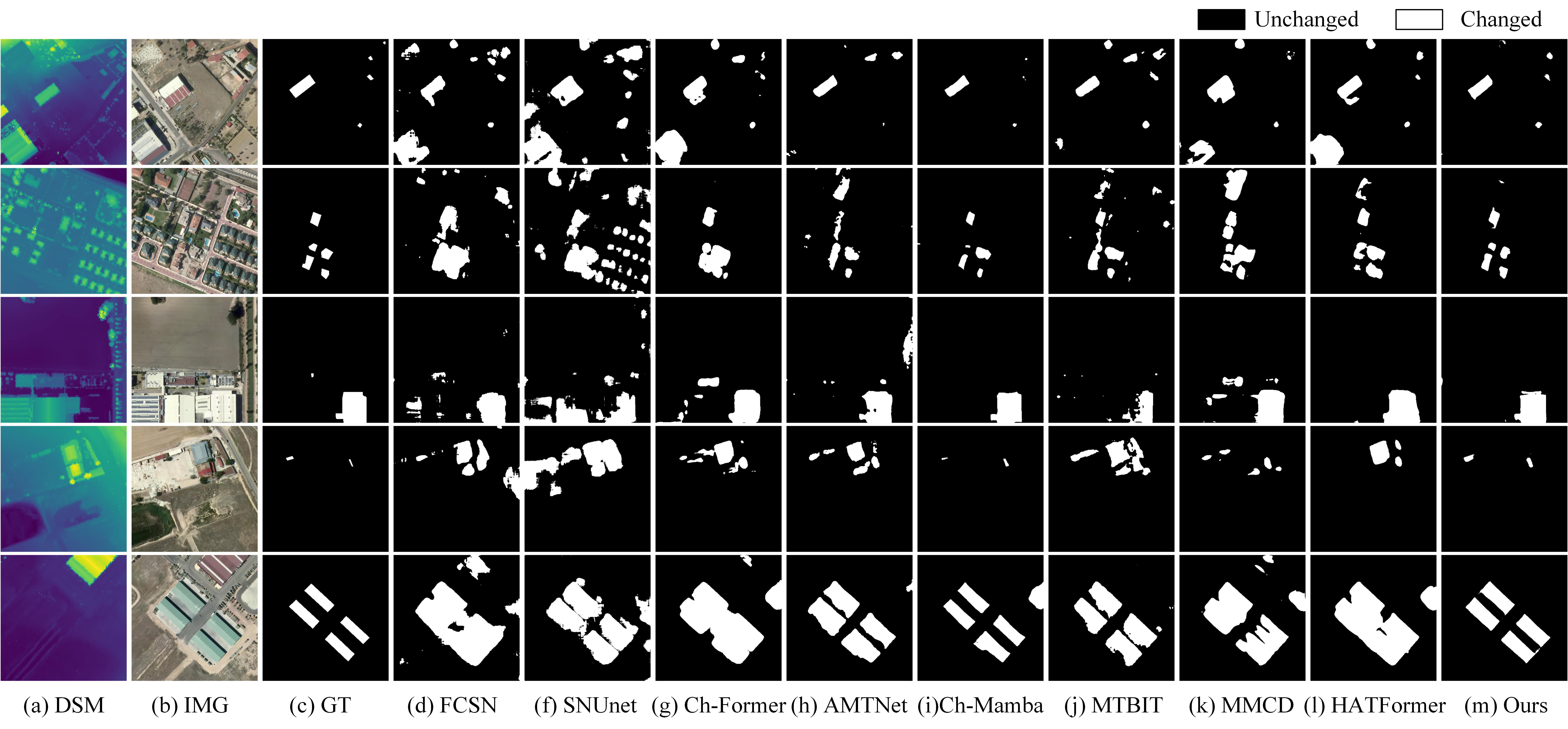}
\caption{Comparison of 2D semantic change detection results of all methods on 3DCD dataset.}
\label{fig:infer3d_2d}
\end{figure*}
\begin{figure*}[]
\centering
\includegraphics[width=\linewidth]{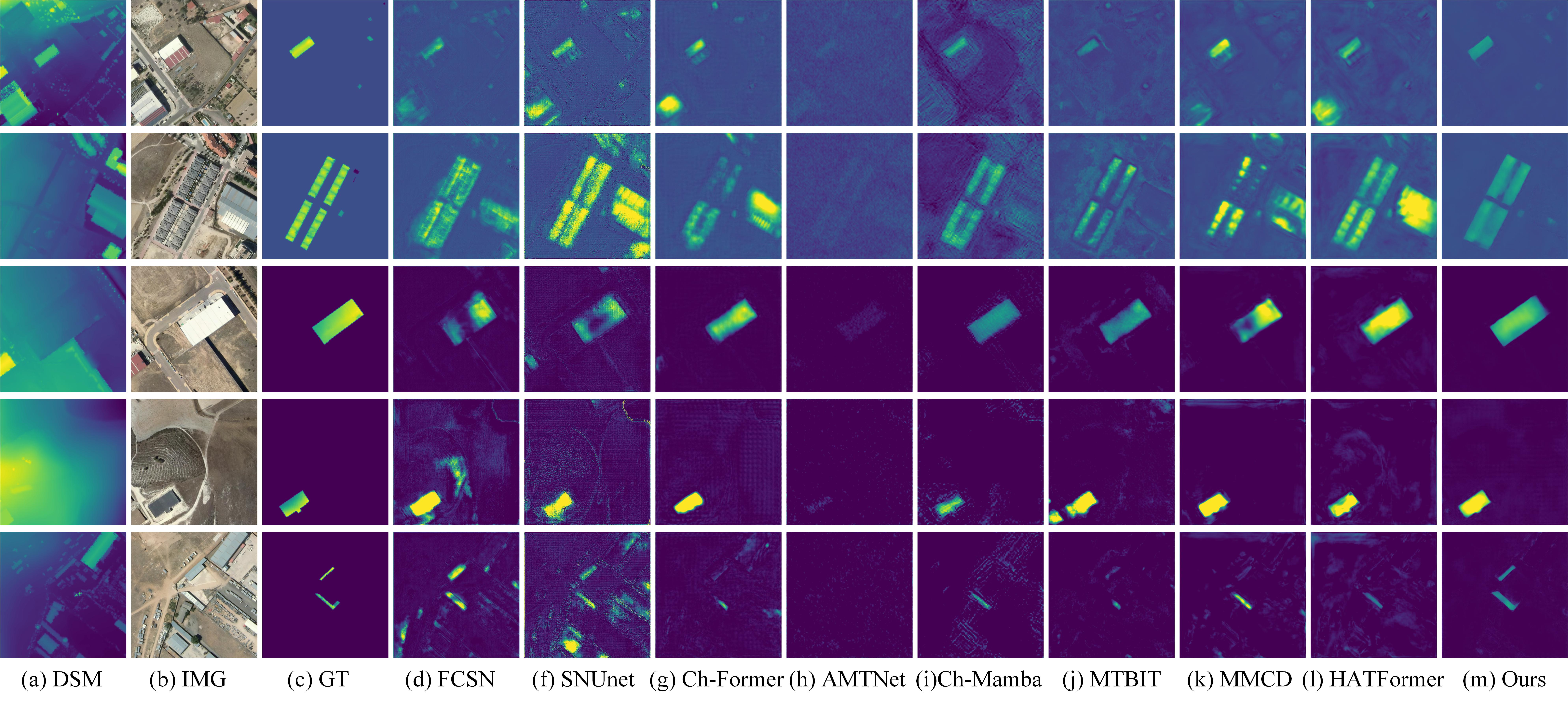}
\caption{Comparison of 3D height change detection results of all methods on 3DCD dataset.}
\label{fig:infer3d_3d}
\end{figure*}
\begin{figure}[]
\centering
\includegraphics[width=\linewidth]{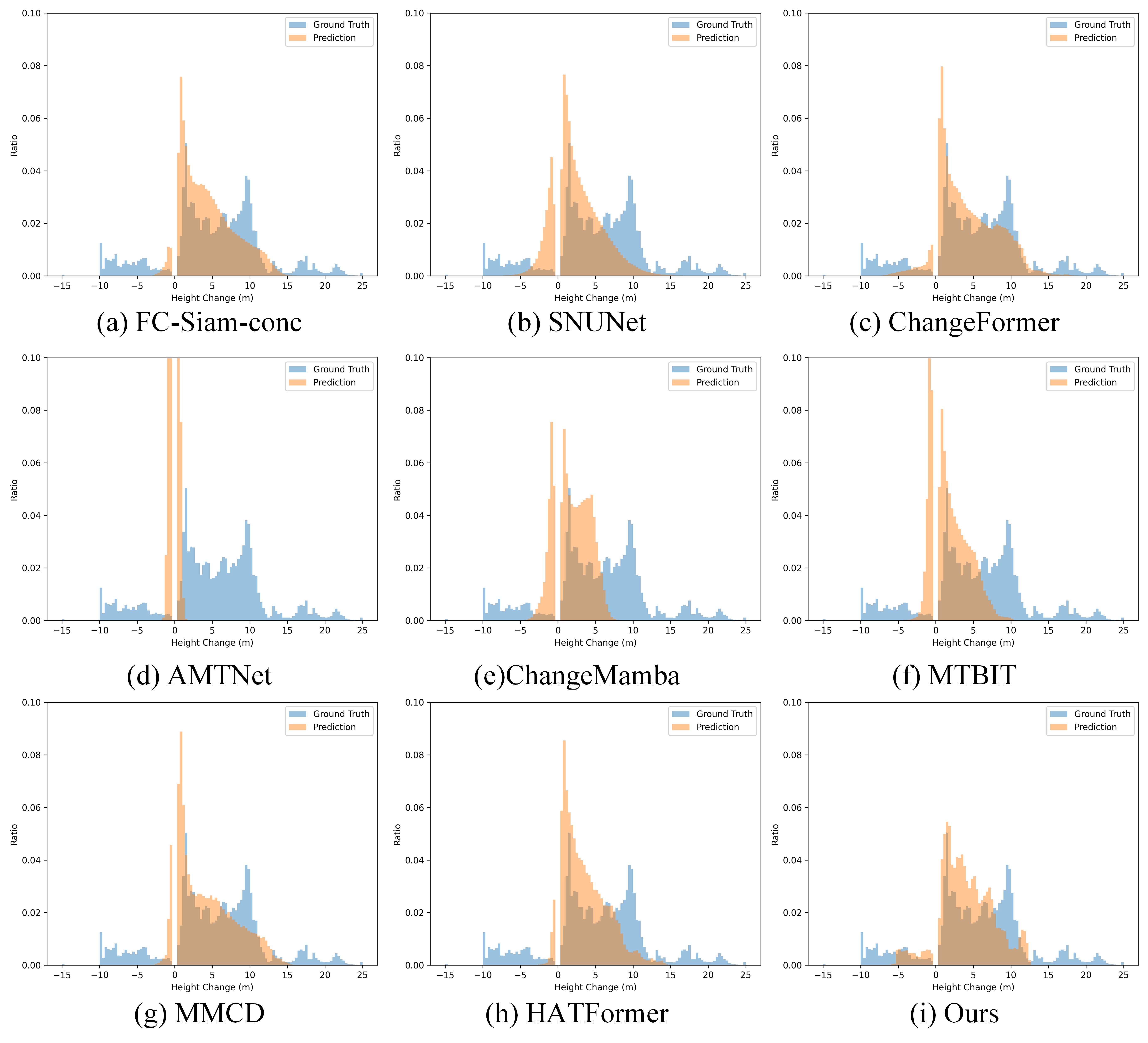}
\caption{Distribution statistics of ground-truth and predicted height changes on the 3DCD test set.}
\label{fig:height-infer3d}
\end{figure}

\begin{table*}[]
\centering
\caption{Quantitative evaluation results on 3DCD dataset.}
\label{tab:infer}
\begin{tabular}{l|cc|cccc}
\hline
\multicolumn{1}{c|}{\multirow{2}{*}{Method}}                & \multicolumn{2}{c|}{2D task metric (\%)} & \multicolumn{4}{c}{3D task metric (m)} \\ \cline{2-7} 
\multicolumn{1}{c|}{}                                       & IoU\_un$\uparrow$              & IoU\_ch $\uparrow$           & MAE$\downarrow$       & RMSE$\downarrow$     & cRMSE$\downarrow$    & cRel$\downarrow$     \\ \hline
FC-Siam-conc\citep{daudt2018fully}         & 90.64               & 21.35       & 0.552   & 1.527   & 5.891 &0.776  \\
SNUNet\citep{fang2021snunet}               & 88.59               & 14.74       & 1.272   & 2.145 &6.302  & 6.302   \\
ChangeFormer\citep{bandara2022transformer} & 94.76               & 29.36    & 0.463   & 1.470    & 5.594  &0.733 \\
AMTNet\citep{liu2023attention}             & 96.20                & 34.50      & 0.562   & 1.630    & 7.364  &1.002 \\
ChangeMamba\citep{chen2024changemamba}     & 97.35               & 38.65        & 1.080    & 1.716   & 6.128  &0.793  \\
MTBIT\citep{marsocci2023inferring}         & 95.42          & 31.02      & 0.443   & 1.480    & 6.350  &0.794  \\
MMCD\citep{liu2024transformer}             & 95.30          & 27.28       & 0.495   & 1.452   & 5.948 &0.795  \\
HATFormrt\citep{liu2025hatformer}          & 95.40                & 29.14    & 0.507   & 1.452   & 6.022  &0.748 \\
Ours                                                        & \textbf{97.53}               & \textbf{47.24}      & \textbf{0.276 } &\textbf{ 1.211}  & \textbf{5.468}  &\textbf{0.665}\\ \hline
\end{tabular}
\end{table*}

Tab. \ref{tab:infer} presents the quantitative evaluation of 2D semantic change and 3D height change prediction results of the proposed method and comparison methods on the 3DCD dataset. In contrast to the multi-class settings of Hi-BCD and NYC-MMCD, the 3DCD dataset is formulated as a binary change detection task, where the primary objective is to distinguish between changed and unchanged regions.

Regarding the binary change detection task, the proposed method also achieves better performance in terms of both accuracy and boundary completeness. As visualized in the fifth row of Fig. \ref{fig:infer3d_2d}, although all methods can generally detect change regions, most comparison methods fail to distinguish the four closely adjacent buildings and incorrectly merge them into a single change region. Conversely, the proposed method preserves building shapes more accurately and captures subtle spatial separations.
Furthermore, as illustrated in the first and second rows, scenes with fragmented building distributions or complex objects tend to cause false detections in unchanged regions. 
%However, the proposed method can effectively identify unchanged areas.

Regarding the 3D task, the proposed method achieves the lowest RMSE and cRMSE, indicating more accurate height change estimation.
The proposed method also demonstrates strong consistency in predicting building height changes. As visualized in the third row of Fig. \ref{fig:infer3d_3d}, for a large-scale individual building, the predicted heights across the rooftop are exhibit a high degree of consistency. In contrast, other methods exhibit significant fluctuations within the building, resulting in noticeable patchy artifacts. Conversely, the proposed method produces a much more uniform color distribution, with values closer to the ground truth.
Moreover, the proposed method shows enhanced sensitivity to fine-grained structures. As illustrated in the fifth row of Fig. \ref{fig:infer3d_3d}, for slender and relatively small-scale buildings, other methods tend to produce incomplete or fragmented detections, whereas the proposed method maintains better height consistency and morphological integrity.

For the 3DCD dataset, we examine the statistical distribution of all ground-truth and predicted height changes in the test set. Since unchanged samples dominate the dataset,we apply a threshold of 0.5 m to the absolute height change values to enhance visual clarity, as illustrated in Fig. \ref{fig:height-infer3d}. 
The distributional characteristics of these histograms provide insightful indicators of the model's predictive fidelity, particularly regarding range coverage and peak alignment. As depicted in Fig. \ref{fig:height-infer3d}, while MMCD and HATFormer exhibit positive height-change peaks that relatively align with the ground truth, they struggle to recover the secondary peak situated around 10 m. Conversely, the proposed method achieves the closest match to the ground-truth peak structures across different height change ranges.
Furthermore, around 0 m, the peak of the proposed method aligns most closely with the ground truth, indicating that the model is highly robust in distinguishing true changes from background noise.

\section{Discussion \label{sec_dicussion}}
\subsection{Ablation study}
\begin{table*}[]
\centering
\caption{Quantitative ablation results of different modules on Hi-BCD dataset.}
\label{tab:ablationhibcd}
\begin{tabular}{llll|ccccc|cccc}
\hline
\multicolumn{4}{c|}{settings} & \multicolumn{5}{c|}{2D task metric (\%)}        & \multicolumn{4}{c}{3D task metric (m)} \\ \hline
EDP  & CCAB  & DSSM  & CCA  & IoU\_un & IoU\_d & IoU\_n & mIoU   & mF1   & MAE & RMSE    & cRMSE      & cRel   \\ \hline
     &       &       &        & 99.22   & 64.25  & 45.38  & 54.82 & 80.09 & 0.243  & 1.205 & 7.395 & 1.493 \\
\ding{51}    &       &       &        & 99.27   & 64.76  & 48.15  & 56.46 & 81.08 & 0.236  & 1.163 & 7.157  & 1.434  \\
\ding{51}    &       & \ding{51}     & \ding{51}      & 99.27   & 63.17  & 48.12  & 55.65 & 80.68 & 0.246  & 1.180 & 7.366 & 1.397 \\
\ding{51}    & \ding{51}     &       &        & 99.30    & \textbf{65.96}  & 50.26  & 58.11  & 82.01 & 0.238  & 1.136 & 7.124 & \textbf{1.326} \\
\ding{51}    & \ding{51}     & \ding{51}     &        & 99.33   & 65.75  & 51.30   & 58.53 & 82.27 & 0.241  & 1.107 & 7.124 & 1.379  \\
\ding{51}    & \ding{51}     & \ding{51}     & \ding{51}      & \textbf{99.34}   & 64.10   & \textbf{54.64}  & \textbf{59.37}  &\textbf{ 82.82 }& \textbf{0.225}  &\textbf{ 1.104} &\textbf{ 6.831} & 1.425 \\ \hline
\end{tabular}
\end{table*}

\begin{table*}[]
\centering
\caption{Quantitative ablation results of different modules on NYC-MMCD dataset.}
\label{tab:ablationnyc}
\begin{tabular}{llll|ccccc|cccc}
\hline
\multicolumn{4}{c|}{settings} & \multicolumn{5}{c|}{2D task metric (\%)}        & \multicolumn{4}{c}{3D task metric (m)} \\ \hline
EDP  & CCAB  & DSSM  & CCA   & IoU\_un & IoU\_n & IoU\_d & mIoU   & mF1  & MAE & RMSE    & cRMSE      & cRel   \\ \hline
     &       &       &        & 98.41   & 41.17  & 31.2   & 36.19 & 68.37 & 0.261  & 2.032 & 9.871 & 1.597 \\
\ding{51}    &       &       &        & 98.41   & 39.68  & 31.68  & 35.68  & 68.04 & \textbf{0.260}    & 1.971 & 9.541 & 1.543 \\
\ding{51}    &       & \ding{51}     & \ding{51}      & 98.42   & 43.46  & 32.48  & 37.97  & 69.61 & 0.292  & 1.970 & 9.557 & 1.610 \\
\ding{51}    & \ding{51}     &       &        & 98.40   & 40.65  & 32.66  & 36.66 & 68.75 & 0.264  & 1.915 & 9.344 & 1.622 \\
\ding{51}    & \ding{51}     &\ding{51}     &        & 98.40    & 42.51  & 32.23  & 37.37  & 69.20  & 0.282  & 2.003 & 9.752  & \textbf{1.540 }\\
\ding{51}    & \ding{51}     & \ding{51}     & \ding{51}      & \textbf{98.43}   & \textbf{44.10}   & \textbf{33.61}  & \textbf{38.86}  & \textbf{70.25} & 0.276  & \textbf{1.908} & \textbf{9.344} & 1.576 \\ \hline
\end{tabular}
\end{table*}

\begin{table*}[]
\centering
\caption{Quantitative ablation results of different modules on 3DCD dataset.}
\label{tab:ablationinfer}
\begin{tabular}{llll|ccc|cccc}
\hline
\multicolumn{4}{c|}{settings}                                                                     & \multicolumn{3}{c|}{2D task metric (\%)} & \multicolumn{4}{c}{3D task metric (m)} \\ \hline
EDP  & CCAB  & DSSM  & CCA        & IoU\_un     & IoU\_ch     &mF1   & MAE     & RMSE    & cRMSE    & cRel   \\ \hline
&    &    &    & 96.90        & 38.23      & 76.87    & 0.317  & 1.332 & 6.110 & 0.747 \\
\ding{51} &                        &                        &                        & 96.78       & 41.31      & 78.41    & 0.319  & 1.310 & 5.936 & 0.720 \\
\ding{51} &                        & \ding{51} & \ding{51} & 97.23       & 44.83      & 80.25    & 0.288  & 1.260 & 5.690 &0.663  \\
\ding{51} & \ding{51} &                        &                        & 97.01       & 39.88      & 77.75    & 0.338  & 1.311 & 5.842 & 0.687 \\
\ding{51} & \ding{51} & \ding{51} &                        & 96.65       & 41.71      & 78.58    & 0.327  & 1.282 & 5.691 & 0.668 \\
\ding{51} & \ding{51} & \ding{51} & \ding{51} & \textbf{97.53}       & \textbf{47.24}      &\textbf{ 81.46 }   & \textbf{0.276}  &\textbf{ 1.211} &\textbf{ 5.468} & \textbf{0.665}  \\ \hline
\end{tabular}
\end{table*}
To validate the effectiveness of each component in the proposed method, ablation studies are conducted to further analyze their individual contributions. The ablation study evaluates three core components: the estimated depth prior (EDP) features, the CCAB (Convolutional Channel Attention Block), and the HCFEB (Hierarchical Change Feature Extraction Block). Specifically, the HCFEB is further analyzed by ablating its internal DSSM (Difference-Aware State Space Mixer) and CCA (Cross-Channel Attention) layers.
Ablation experiments are conducted on all three datasets. The results on Hi-BCD, NYC-MMCD, and 3DCD are reported in Tab. \ref{tab:ablationhibcd}, Tab. \ref{tab:ablationnyc}, and Tab. \ref{tab:ablationinfer}, respectively. The outcomes across the three datasets exhibit consistent trends.

First, introducing the estimated depth prior consistently improves both 2D and 3D change detection performance, demonstrating the effectiveness of geometric priors. The depth prior provides a consistency constraint similar to the DSM structure, helping to mitigate the effects of shadows, occlusions, and noise in high-resolution aerial imagery.

The DSSM module plays a pivotal role in change modeling by simultaneously encoding spatio-temporal dynamics across multi-temporal features and capturing intricate change-aware interactions. Its integration significantly improves change-related metrics across all datasets and effectively captures temporal differences between cross-modal inputs.
Furthermore, incorporating the CCA after DSSM leads to consistent and notable performance gains across all datasets, particularly in 2D change detection. This indicates that CCA further augments the discriminative power of change-related feature representations. The combination of DSSM and CCA enables effective local–global change modeling.
Ultimately, the full integration of all proposed modules attains unrivaled performance across all datasets, confirming the mutual reinforcement and robust complementarity of the constituent architectural elements.

\begin{table*}[]
\centering
\caption{Sensitivity analysis of loss weights on different datasets.}
\label{tab:loss}
\resizebox{2\columnwidth}{!}{%
\begin{tabular}{llll|cccc|cccc|cccc}
\hline
\multicolumn{4}{c|}{settings} & \multicolumn{4}{c|}{Hi-BCD dataset}    & \multicolumn{4}{c|}{NYC-MMCD dataset}    & \multicolumn{4}{c}{3DCD dataset}     \\ \hline
$\mathcal{L}_{\mathrm{wCE}}$  & $\mathcal{L}_{\mathrm{mse}}^{3d}$   & $\mathcal{L}_{\mathrm{grad}}$    & $\mathcal{L}_{\mathrm{mse}}^{dsm}$   & IoU\_d & IoU\_n & MAE    & cRMSE  & IoU\_n & IoU\_d & MAE    & cRMSE  & IoU\_un & IoU\_ch & MAE    & cRMSE  \\ \hline
1                            & 1                                 & 0.2                           & 1                                 & 64.10   & \textbf{54.64}  & 0.225 & \textbf{6.831} & 44.11   & \textbf{33.62}   & 0.276  & 9.344 & 97.53  & 47.24  & 0.276 & \textbf{5.468 }  \\
1                            & 1                                 & -                             & 1                                 & 65.84  & 48.59  & 0.325  & 7.449  & 43.59   & 33.47   & 0.493  &\textbf{ 8.865} & \textbf{97.78 } & \textbf{49.16}  & 0.254  & 5.778    \\
1                            & 1                                 & 0.2                           & -                                 & 64.80   & 52.28  & 0.226  & 6.784 & 43.04   & 31.42   & 0.286  & 9.745 & 97.48  & 44.40   & 0.284  & 5.603     \\
1                            & 1                                 & -                             & -                                 & 64.95  & 53.11  & 0.304  & 7.311 & 42.29	   & 32.94    & 0.564 & 8.931  & 97.62  & 45.00     & \textbf{0.249 } & 5.560    \\
1                            & 1                                 & 0.2                           & 0.5                               & \textbf{66.79}  & 51.49  & 0.260   & 7.084 & 43.85   & 33.05   & 0.304  & 9.301  & 97.62  & 47.76  & 0.346  & 5.736   \\
1                            & 2                                 & 0.2                           & 1                                 & 63.20   & 51.07  & 0.248  & 6.872 & \textbf{44.21}   & 32.82   & 0.358  & 9.106  & 97.55  & 48.80   & 0.264  & 5.298    \\
1                            & 1                                 & 0.1                           & 1                                 & 66.33  & 53.94  & 0.254  & 7.132  & 43.15   & 32.34   & 0.322  & 9.157 & 97.40   & 46.64  & 0.270   & 5.622   \\
1                            & 1                                 & 0.3                           & 1                                 & 66.04  & 48.90   & \textbf{0.224}  & 7.249  & 42.95   & 30.97   & \textbf{0.269} & 9.803 & 97.44  & 45.43  & 0.320   & 5.838 
 \\ \hline
\end{tabular}%
}
\end{table*}

\subsection{Parameters Sensitivity Analysis}
The overall loss function of the proposed network consists of a weighted cross-entropy loss for 2D semantic change $\mathcal{L}_{\mathrm{wCE}} $, an MSE loss for 3D height change prediction $\mathcal{L}_{\mathrm{mse}}^{3d}$, an auxiliary DSM MSE loss 
$\mathcal{L}_{\mathrm{mse}}^{dsm}$, and a gradient loss for 3D height changes in change regions 
$\mathcal{L}_{\mathrm{grad}}$.

As defined in Eq. \ref{eq:allloss}, various loss terms are assigned different weights.
To assess the impact of these hyperparameters on model performance, we further conduct parameter sensitivity experiments on the weights of these loss terms across the three datasets.The resulting quantitative performance under different weight configurations is summarized in Tab. \ref{tab:loss}

The results indicate that individual loss terms have different impacts on 2D semantic change segmentation and 3D height change regression. Specifically, the incorporation of moderate gradient loss and DSM loss effectively improves the overall performance. Increasing the weight of the 3D loss slightly enhances height regression metrics but leads to a decline in 2D semantic segmentation performance.

The introduction of gradient loss enhances the precision of height change metrics by strengthening constraints on height boundaries and local structures. However, when its weight is too large, excessive gradient constraints may amplify local noise and affect overall optimization stability.
Furthermore, adopting DSM prediction corresponding to T2 imagery as an auxiliary task not only benefits 2D change detection but also indirectly improves 3D change modeling. The geometric prior provided by DSM facilitates the identification of semantic change regions, thereby enhancing the joint optimization of segmentation and regression tasks.

Finally, the first set of parameters is selected as the final configuration, as it demonstrates better cross-dataset robustness while maintaining a balanced performance between 2D semantic change segmentation and 3D height change regression.
\section{Conclusion \label{sec_conclusion}}
In this paper, we investigate the challenging cross-modal change detection task utilizing pre-event DSM and post-event imagery, proposing a unified framework for joint 2D semantic change detection and 3D height change prediction. To facilitate research in complex urban environments, we introduce NYC-MMCD, a novel multi-modal and multi-task change detection dataset.

To bridge the substantial modality gap between imagery and DSM, the proposed method introduces estimated depth as a geometric prior for the image modality, thereby strengthening the cross-modal correlation. An adaptive gated fusion mechanism is further designed to integrate this depth prior while preserving the dominance of spectral-geometric features, ensuring superior feature alignment. Building upon this, a multi-stage cross-temporal cross-modal fusion architecture is developed to hierarchically extract inter-modal discrepancies, followed by a multi-task decoding framework that simultaneously models 2D and 3D changes. Furthermore, DSM auxiliary supervision is incorporated to enforce structural consistency and provide robust geometric constraints for high-precision 3D regression. Extensive experiments on three benchmarks demonstrate the superiority of the proposed approach over state-of-the-art methods.

Compared with conventional paradigms that necessitate complete bi-temporal 3D acquisitions, the proposed DPG-CD framework facilitates the simultaneous extraction of 2D and 3D building dynamics utilizing only pre-event DSM and post-event imagery. By alleviating the dependency on multi-temporal 3D data, our approach offers superior operational scalability for large-scale and high-frequency monitoring tasks. Consequently, this methodology holds immense promise for diverse real-world applications, ranging from autonomous urban surveillance and rapid disaster damage assessment to precision urban management.
\nolinenumbers

\section*{CRediT authorship contribution statement}
\textbf{Luqi Zhang:} Writing - original draft, Methodology, Data curation.  \textbf{Zhen Dong:} Writing - review \& editing, Supervision.   \textbf{Bisheng Yang:} Writing - review \& editing, Supervision, Funding acquisition.

\section*{Acknowledgment}
This study was jointly supported by the National Natural Science Foundation Project (No.42130105).

%% Loading bibliography style file
% \bibliographystyle{cas-ref}
% \bibliographystyle{cas-model2-names}

% Loading bibliography database

%\bibliographystyle{cas-model2-names}
\bibliographystyle{unsrtnat}
\bibliography{cas-refs}

\end{document}